\documentclass[11pt]{article}
\usepackage{amsmath}

\usepackage[preprint]{acl}
\usepackage{enumitem}

\usepackage{times}
\usepackage{latexsym}

\usepackage[T1]{fontenc}

\usepackage[utf8]{inputenc}

\usepackage{microtype}

\usepackage{inconsolata}

\usepackage{graphicx}

\usepackage{booktabs}
\usepackage{multirow}

\usepackage{subcaption}
\usepackage{caption}

\usepackage[most]{tcolorbox}
\tcbuselibrary{breakable} 

\newtcolorbox{promptbox}[1][]{
  enhanced,
  colback=gray!5,         
  colframe=gray!40,       
  boxrule=0.5pt,          
  arc=4pt,                
  fonttitle=\bfseries,
  coltitle=black,
  attach boxed title to top left={xshift=10pt, yshift=-10pt},
  top=15pt,               
  title={#1}
}

\title{User Feedback Provides a Unique Signal that LLMs Can not Detect}

\author{
 \textbf{Shachar Don-Yehiya\textsuperscript{1,2}} \qquad
 \textbf{Leshem Choshen \textsuperscript{2,3,4}} \qquad
 \textbf{Omri Abend\textsuperscript{1}} \\
 \textsuperscript{1}The Hebrew University of Jerusalem,
 \textsuperscript{2}IBM Research,
 \textsuperscript{3}MIT,
 \textsuperscript{4}MIT-IBM Watson AI Lab \\
   \texttt{\{first.last\}@mail.huji.ac.il}
}

\begin{document}

\maketitle
\begin{abstract}
Harnessing naturally occurring feedback from user interactions offers a promising learning signal for Large Language Models (LLMs). However, recent studies suggest this feedback is inherently noisy and difficult to leverage effectively. We challenge this conception by demonstrating that user feedback is a highly actionable signal for improvement, and that its perceived ineffectiveness stems from a systematic bias in current evaluation paradigms. To isolate the usefulness of feedback, we construct synthetic data with a definitive ground truth, alongside naturalistic data to validate that our findings hold in real-world scenarios. By comparing model revisions generated with and without access to feedback across both settings, we show that feedback-informed revisions resolve targeted issues at significantly higher rates than baseline revisions. Finally, we expose the root of the evaluation bias: when a model successfully fixes an issue exclusively due to feedback, LLM judges frequently fail to identify the genuinely corrected response, systematically preferring inferior baseline outputs instead.
\end{abstract}

\section{Introduction}

As Large Language Models (LLM) are being used in more open-ended tasks without clear success signals, aligning the models with human intent becomes challenging. Inspired by the spontaneous feedback humans provide in natural conversation \citep{VRANJES201815, doi:10.1080/10904018.2010.508675, bassiri2011interactional, 5a61a38b-79fc-30d4-a76e-8a66a7707021, pickering_garrod_2021}, prior research has explored extracting naturally occurring feedback from user interactions, mainly with chat models \citep{naturallyoccurringfeedback, lin-etal-2024-interpretable, shi-etal-2026-wildfeedback}. While naturally occurring feedback has high potential as a learning signal \citep{DonYehiya_2025a, buening2026aligning, jin2025era}, recent work argues that it is inherently noisy and difficult to use effectively \citep{liu-etal-2025-user}.

\begin{figure}[t!]
\includegraphics[width=\linewidth]{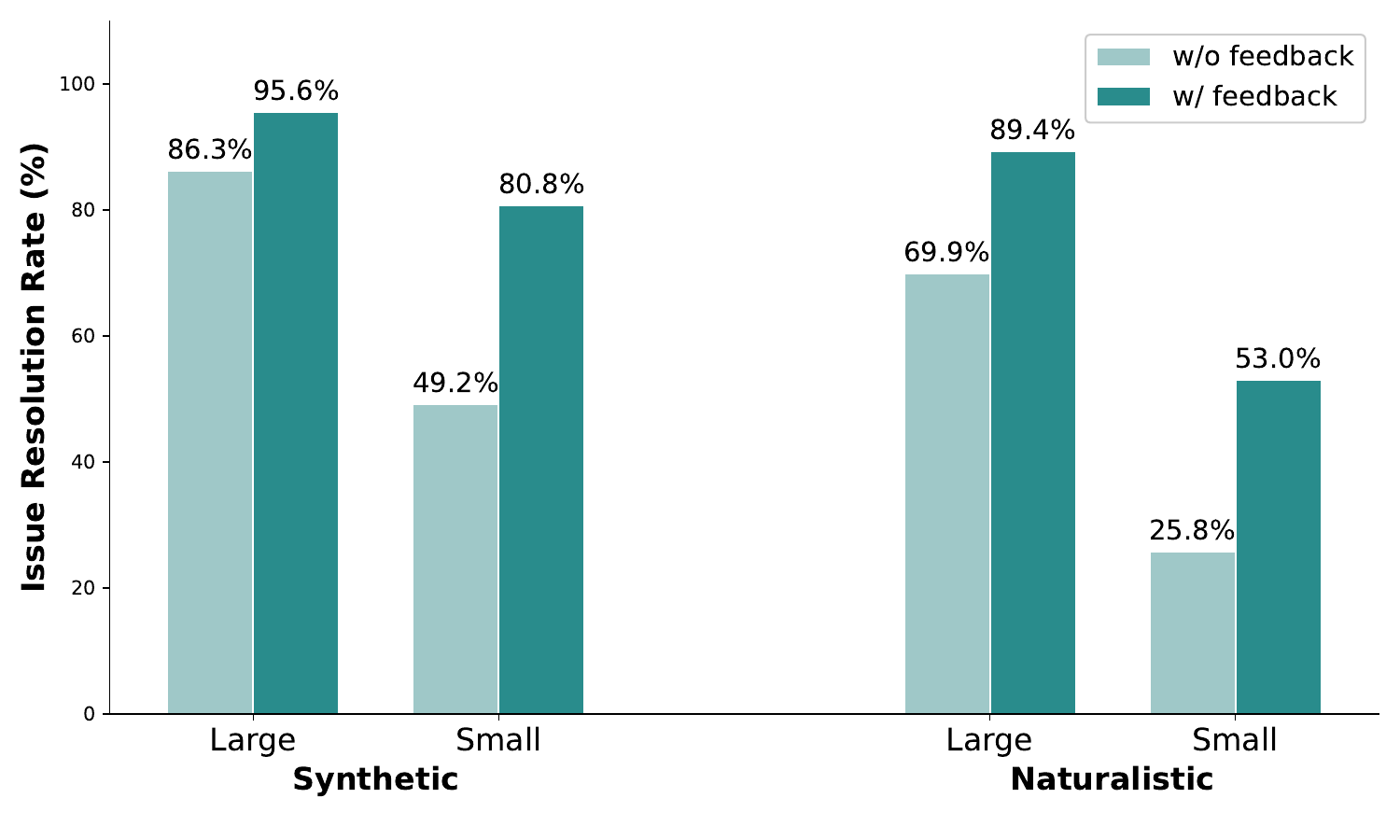}
\caption{Issue resolution rates for large and small improvers on synthetic and naturalistic data. The synthetic data results represent the average across all corruption types. For both settings, providing feedback ("w/ feedback") yields a consistent positive delta over the baseline ("w/o feedback"). While baseline performance drops on the more challenging naturalistic data, the relative benefit of providing feedback remains strong.}
\label{fig:fa_bar_plot}
\end{figure}

This paper challenges this common conception. Through controlled experiments, we find that naturally occurring feedback is not inherently ineffective; on the contrary, it is a strong signal for improvement, but its contribution is masked by evaluation paradigms that are systematically biased against it.

To isolate the utility of feedback, we compare model revisions made with and without feedback on two types of data. We ensure feedback targets real issues and a ground truth is available through synthetic data, where we deliberately corrupt existing model responses (\S\ref{sec:corruption}). In contrast, we validate our findings 
hold true in reality, 
by extracting user feedback from real conversations (\S\ref{sec:real_data_extraction}).\footnote{Code: \url{https://github.com/shachardon/feedback-blindspot} Data: \url{https://huggingface.co/datasets/shachardon/feedback-blindspot}}


Across both settings, we find that the feedback is highly effective (\S\ref{sec:feedback_adrr}). With-feedback improvements resolve the targeted issues at significantly higher rates than their without-feedback baselines—yielding a $9\%-32\%$ increase in resolution on the synthetic data, and a replicated $16\%-27\%$ increase on the real user data. This raises a critical question: if feedback demonstrably drives such improvements, why was it previously found to be unhelpful?

We investigate the evaluation mechanism as the primary source of confusion. We directly compare the two improvement variants using a strong LLM judge \citep{chiang-lee-2023-large, liu-etal-2023-g}. By isolating the instances across our data where the feedback-informed response successfully fixed the issue while the baseline response failed, we uncover a critical vulnerability in current evaluation methods (\S\ref{sec:intersection}). For instances where the model failed to improve without external feedback, the LLM judge frequently fails to identify the genuinely corrected response (\S\ref{sec:models_struggle}). In our naturalistic experiments, for example, the judge correctly preferred the response that solved the problem in only $34\%-54\%$ of the cases. Consequently, the evaluation becomes systematically biased against the feedback signal, penalizing genuine corrections in favor of general stylistic changes (\S\ref{sec:improvement_type}).





\section{Method}

We examine whether user feedback provides a useful signal. 
To simulate a clean scenario where we can easily verify whether the targeted issue was resolved, 
we couple the naturalistic data experiments with synthetic data experiments.     

\subsection{Synthetic Data Compilation} \label{sec:corruption}
To compose our synthetic data, we take tuples of user queries and model responses, and ``corrupt'' the responses to make them invalid. We then examine whether simulated user feedback is useful for recovering from the corruption.  

\subsubsection{Response Corruption}
We define four corruption types:

\begin{itemize}[leftmargin=1em, itemsep=0pt, topsep=0pt]
    \item \textbf{Causality Inversion}: Swaps the cause and effect in a causal relationship from the reference response, producing a fluent but backward claim. 
    \\\textit{Example:} ``The high inflation rate forced the central bank to raise interest rates'' becomes ``The central bank raising interest rates forced the high inflation rate.''
    
    \item \textbf{Crucial Omission}: Deletes a detail that is essential for the response to be correct, safe, or functional, smoothing over the surrounding text so the gap is not immediately obvious. 
    \\\textit{Example:} ``Execute \texttt{DELETE FROM users WHERE status = 'inactive'} to clear out the old accounts'' becomes ``Execute \texttt{DELETE FROM users} to clear out the old accounts.''
    
    \item \textbf{Entity/Subject Swap}: Replaces a key entity (e.g., a person, location, number, library, or variable name) with a related but factually incorrect one, while keeping the sentence structure and tone unchanged. 
    \\\textit{Example:} ``Import the \texttt{pandas} library to manipulate the dataframe'' becomes ``Import the \texttt{numpy} library to manipulate the dataframe.''
    
    \item \textbf{Logic Operator Reversal}: Inverts a logical relationship in the response, such as a comparison operator, boolean value, conditional, or chronological order, so the text or code still reads naturally but the underlying logic is flipped. 
    \\\textit{Example:} ``Proceed with the installation only if \texttt{is\_admin == True}'' becomes ``Proceed with the installation only if \texttt{is\_admin == False}.''
\end{itemize}

\paragraph{}Given the user query $q$, the original model response $r$ and a corruption type, we prompt a model $M_c$ (henceforth, the {\it corrupting model}) to corrupt the model response according to the corruption type, making minimal changes. The corrupting model is asked to provide a new corrupted response $r_c$, a change log $c_{log}$ explaining the corruption, a difficulty score $s_d \in \{LOW, MEDIUM, HIGH\}$ to indicate how hard it is for a generic reader to spot the error, and three levels ``user feedback'' $\{f_{sol}, f_{prob}, f_{binary}\}$, the first level telling how to fix, the second only pointing out the problem, and the third only saying that the model is wrong (see \S\ref{app:corruptions_prompt} for the full prompts).
Formally, we have
$$M_c(q,r,c\_type) = \left(r_c, c_{log}, f_{sol}, f_{prob}, f_{binary}\right)$$

\subsection{Naturalistic Data Extraction} \label{sec:real_data_extraction}

To collect real user feedback data, we follow the feedback extraction method of \citet{naturallyoccurringfeedback}. We prompt \textit{Qwen3-30B-A3B-Instruct-2507} to recognize spans of naturally occurring feedback and classify them into a taxonomy. The taxonomy contains $4$ negative categories (\textit{Repeat or Rephrase}, \textit{Make Aware with Correction}, \textit{Make Aware without Correction}, \textit{Ask for Clarification}) and one positive (\textit{Positive Feedback}). We filter out the positive feedback, as in our experiments we focus on responses improvements (see Appendix~\ref{app:nof_extraction}).

Manually examining a sample of extracted feedback examples, we find that some instances that have been recognized as feedback, contain feedback that is subjective. Although these feedback instances might be useful for personalization, they should not be used for general alignment training. For example, a user asks for a pasta recipe, and then later provides feedback saying that it should be vegan (\textit{Make Aware with Correction}).

To overcome this, we prompt a model, \textit{Gemini-3-Flash}, to decide whether a given feedback contains actionable insights that should apply to all future users asking the same question or not, see Appendix~\ref{app:nof_extraction} for the full prompt.

\subsection{Improvement with and without Feedback} \label{sec:improvement}
Following \citet{liu-etal-2025-user}, we prompt a model $M_{i}$, the ``improver'', to improve the corrupted response with and without feedback $f \in \{f_{sol}, f_{prob}, f_{binary}\}$, for simplicity we denote it with $f$. We end up with two improved responses, $M_i(q,r_c,f)=r_{+f}$ and $M_i(q,r_c) = r_{-f}$ accordingly. See \S\ref{app:improvement_prompts} for the full prompts. 


\section{Evaluation} \label{sec:eval}

\subsection{Issue Resolution Evaluation} \label{sec:feedback_adrr_eval}
To examine whether the improved responses $r_{+f}$ and $r_{-f}$ indeed recovered the corruption, we prompt a judge model $J_{IR}$ to assess whether the improved responses addressed $f_{sol}$, i.e., whether they addressed the simulated feedback that explains how to fix the issue. Thus, we confirm that the improved responses fixed the corruption correctly.
The judge is provided with $q, r_c,f_{sol}$ and an improved response ($r_{+f}$ or $r_{-f}$). The judge is instructed to summarize the feedback intent, compare the changes between the corrupted response and the improved one, check for regression, and finally provide a score (1 is poor 5 is perfect) and a binary decision whether the improved response addressed the feedback or not (see Appendix~\ref{app:prompt_eval}).
We use this measure also for the naturalistic data. Note that in this case, we do not have the guaranty that addressing the feedback would solve the issue. For example, \textit{Make Aware without Correction} feedback will not provide us with the sufficient information to verify the revision correction. We therefore use this for the naturalistic data as an approximation only.

\subsection{Pairwise Evaluation} \label{sec:eval_pairwise}
Given the user query $q$, we use a judge model $J_{P}$ to compare two responses directly, response $A$ and response $B$, to determine which response is the best. For example, we compare the original response and the corrupted one $J_P(q,r,r_c)$, or the two improvements variants $J_P(q,r_{+f},r_{-f})$. The judge outputs a reasoning explaining its choice, and a verdict $\in \{A, B, Tie\}$. To avoid biases we set the order of the responses randomly. See Appendix~\ref{app:prompt_eval} for the full evaluation prompt.

\section{Experimental Setup} \label{sec:setup}
\paragraph{Data.} For the synthetic setting, we use \textit{Arena-Hard-v2.0}  \citep{li2024crowdsourced}. This dataset contains a \textit{hard prompt} category, comprising 500 challenging real-world user queries (open-ended software engineering problems, math questions, logic puzzles, etc.), and a \textit{creative writing} category, comprising 250 creative writing queries sourced from Chatbot Arena \citep{chiang2024chatbot}. 
We use the model responses of the \textit{o3} model, which is considered to be strong, as we assume that prior to the corruption the response was valid.
For naturalistic experiments, we use data from the ShareLM collection \citep{don-yehiya-sharelm}, a collection of open human-model interaction datasets. We filter for English conversations only, for annotations reasons. 

\paragraph{Models.} For $M_c$ we use \textit{Gemini-3-flash-preview} \citep{team2023gemini}. For the \textbf{improver model} we experiment with two models, differing in size: we use \textit{Gemini-3-flash-preview}, marked as $M_i^{large}$, and \textit{Qwen3-8B} \citep{yang2025qwen3}, marked as $M_i^{small}$. For the \textbf{judge model} we use a stronger model, \textit{Gemini-3.1-Pro-preview}. As detailed in \S\ref{app:budget}, the multi-stage inference required for corruption generation, improvement, and evaluation incurs substantial computational expense. We therefore restrict our evaluation to this representative set of models. However, we report initial results for \textit{GPT-OSS-20B} in Appendix~\ref{app:gpt_oss}, demonstrating consistent trends across architectures.


\begin{table*}[t!]
\centering
\small 
\setlength{\tabcolsep}{4pt} 
\begin{tabular}{llccccc}
\toprule
\textbf{Improver} & \textbf{Judge} & \textbf{Causality} & \textbf{Crucial} & \textbf{Entity} & \textbf{Operator} & \textbf{Average} \\
& \textbf{Preference} & \textbf{Inversion} & \textbf{Omission} & \textbf{Swap} & \textbf{Reversal} & \textbf{} \\
\midrule
\multirow{3}{*}{\textbf{Large}} 
& w. Feedback & 28.5\% & 28.7\% & 25.9\% & 25.8\% & 27.2\% \\
& w/o. Feedback & \textbf{36.9\%} & \textbf{37.5\%} & \textbf{38.7\%} & \textbf{37.9\%} & \textbf{37.8\%} \\
& Tie & 34.6\% & 33.8\% & 35.4\% & 36.3\% & 35.0\% \\
\midrule
\multirow{3}{*}{\textbf{Small}} 
& w. Feedback & \textbf{45.7\%} & \textbf{50.0\%} & \textbf{52.4\%} & \textbf{56.4\%} & \textbf{51.2\%} \\
& w/o. Feedback & 39.2\% & 37.4\% & 30.0\% & 31.2\% & 34.3\% \\
& Tie & 15.1\% & 12.6\% & 17.7\% & 12.5\% & 14.5\% \\
\bottomrule
\end{tabular}
\caption{Judge pairwise evaluation of the with-feedback vs. without-Feedback variants. Although the without-feedback variant fix the corruption in a lower rate, we see a consistent preference for the without-feedback variant for the large improver, across the four corruption types.}
\label{tab:with_feedback_vs_whithout_feedback}
\end{table*}

\begin{figure*}[t!]
    \centering
    \begin{subfigure}[b]{0.45\textwidth}
        \centering
        \includegraphics[width=\linewidth]{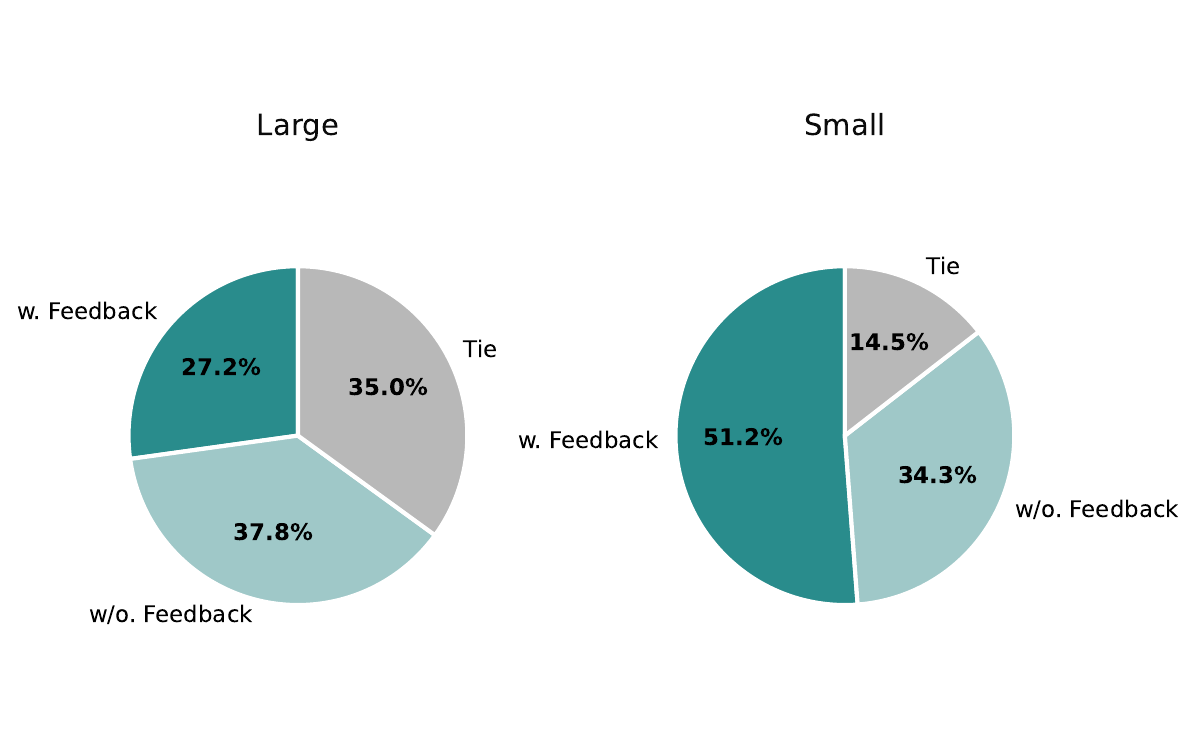}
        \caption{Synthetic}
        \label{fig:synthetic_pie_plot}
    \end{subfigure}
    \hfill
    \begin{subfigure}[b]{0.45\textwidth}
        \centering
        \includegraphics[width=\linewidth]{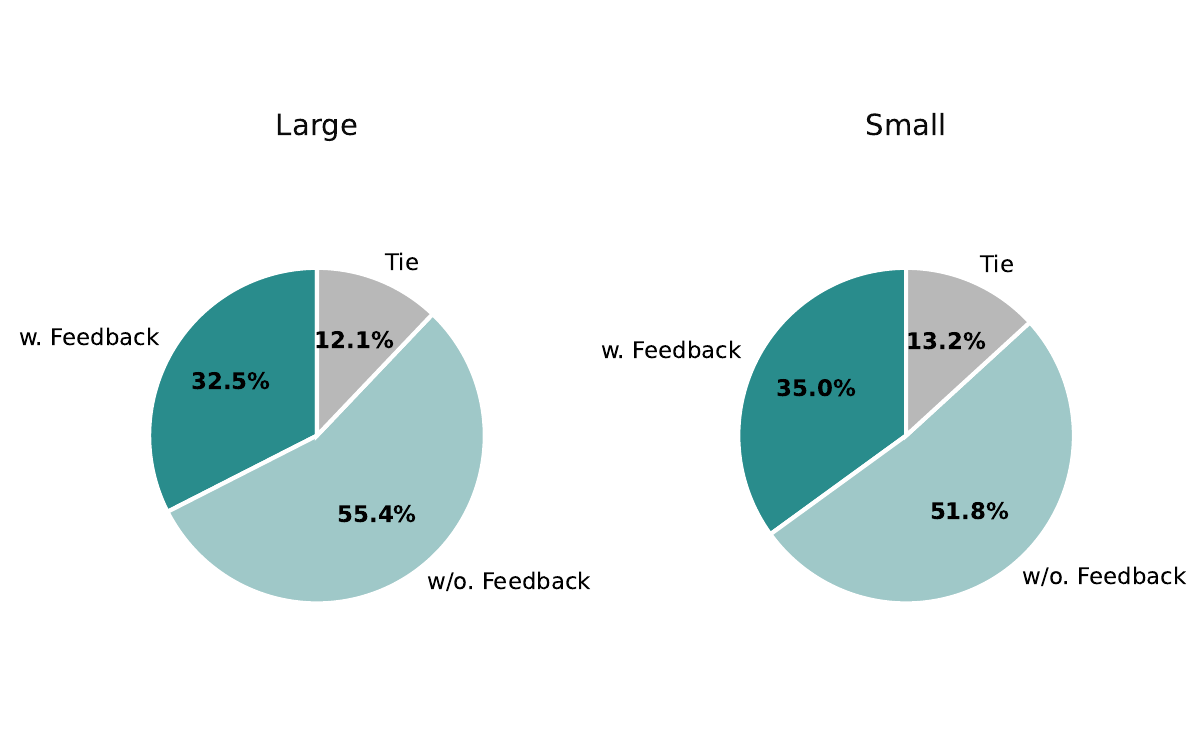}
        \caption{Naturalistic}
        \label{fig:naturalistic_pie_plot}
    \end{subfigure}

    \caption{Judge pairwise preference evaluation for with-feedback vs. without-feedback variants.
    Notably, judges frequently prefer the without-feedback variants despite those variants fixing corruptions at a lower rate.}
    \label{fig:pairwise_pies}
\end{figure*}


\paragraph{Corruption Verification.} \label{sec:corruption_verification}

To verify that the corruption process successfully rendered the new responses invalid, we manually evaluated $96$ corrupted samples: $64$ samples from the hard prompt category and $32$ from the creative writing category. 
See Appendix \ref{app:human_annotation} for the detailed annotation guidelines and procedures. 
Out of the hard prompt samples, $92.2\%$ were marked as correct (indicating the corrupted response was successfully invalidated), $3.1\%$ as borderline, $1.6\%$ as incorrect, and $3.1\%$ as unfamiliar (where annotators lacked the domain expertise to evaluate the question). For the creative writing category, $59.4\%$ of the samples were marked as correct, and the rest $40.6\%$ as incorrect.
Following these results, to ensure a clean evaluation, we report results exclusively for the hard prompt category in the main text unless otherwise specified. The comprehensive results, including the creative writing category, are provided in Appendix~\ref{app:full_results}.

\paragraph{Feedback Relevance Verification.}
Filtering the naturalistic data to include only feedback cases that should be applied to all future users (see \S\ref{sec:real_data_extraction}), the model deemed $47.9\%$ of the naturally occurring feedback cases as relevant. 
To verify the model's decisions, we ask a human annotator to review $165$ samples of model verdicts and reasoning, see Appendix \ref{app:human_annotation} for the full prompt.
$86.1\%$ of the verdicts were marked as correct (the model was right to tag the feedback as relevant/irrelevant), $7.3\%$ as incorrect, and $6.7\%$ as unfamiliar.

\paragraph{Issue Resolution Verification.}
To validate our issue resolution evaluation method, we manually annotated $110$ samples from the synthetic data (hard prompt category only) and $65$ samples from the naturalistic data.
For the synthetic data, excluding $16$ samples that were marked as unfamiliar, the remaining annotations showed Cohen's Kappa of $0.81$.
For the naturalistic data, excluding $2$ samples that were marked unfamiliar, we get Cohen's Kappa of $0.24$.
See Appendix \ref{app:human_annotation} for the full guidelines.

\section{Results}
We generate $r_{+f}$ and $r_{-f}$ for 2000 ($500$ hard prompt samples $\times 4$ corruption types) synthetic samples, and 1000 naturalistic data samples (\S\ref{sec:improvement}).

\subsection{The Value of External Feedback} \label{sec:feedback_adrr}

To evaluate the overall impact of external feedback, we use the issue resolution evaluation (\S\ref{sec:feedback_adrr_eval}) to assess whether the improved responses successfully solved the issues. We compare the with-feedback variant, $J_{FA}(q,r_c,r_{+f},f)$, against the without-feedback variant, $J_{FA}(q,r_c,r_{-f},f)$. 

Figure~\ref{fig:fa_bar_plot} presents the issue resolution rates for both synthetic and naturalistic data. 
We find that \textbf{external feedback consistently helps fix and improve responses, successfully addressing issues in 9\%--35\% more cases than when feedback is withheld.} Furthermore, while smaller models resolve issues independently at a much lower rate, they benefit far more substantially from the inclusion of feedback than larger models.

In the synthetic setting, the large improver successfully incorporated feedback 95.6\% of the time (compared to 86.3\% without feedback), while the small improver achieved 77.9\% (compared to just 42.7\% without). The naturalistic data confirms these trends but, as expected, proves to be a more challenging and noisy environment. Without feedback, the models independently resolved only 70\% (large) and 26\% (small) of the cases. With feedback, their issue resolution rates jumped to 89\% and 53\%, respectively. While the absolute issue resolution rates are lower in the naturalistic setting compared to the synthetic one, the performance gain provided by the feedback remains robust.

\subsection{The Pairwise Evaluation Discrepancy} \label{sec:results_pairwise}
Following previous works that sought to quantify the benefit of naturally occurring feedback \citep{liu-etal-2025-user, jin2025era}, we compare the feedback-enhanced responses against the without-feedback baseline. 

We conduct an LLM-as-a-Judge pairwise comparison between the two improved responses: $r_{+f}$ and $r_{-f}$ (\S\ref{sec:eval_pairwise}). Fig.~\ref{fig:pairwise_pies} presents the results for both the synthetic and naturalistic data. We find that \textbf{although the without-feedback variants resolve the targeted issues at a lower rate, in most settings the pairwise judge consistently prefers them over their with-feedback counterparts,} in line with previous works finding.

For the synthetic data, the judge prefers the without-feedback improvements of the large improver over the with-feedback improvements. This trend is reversed for the small improver, where the with-feedback improvements prove superior. We address this disparity in \S\ref{sec:self_judge}. Nevertheless, Table~\ref{tab:with_feedback_vs_whithout_feedback} shows that these results are consistent across all four corruption types. 

For the naturalistic data, the judge prefers the without-feedback variants generated by both the large and small improvers.


Importantly, this reveals more than a mere lack of improvement due to the feedback. If the improver simply failed to benefit from the feedback, we would expect a balanced judge to be indifferent between $r_{+f}$ and $r_{-f}$. Instead, we find that responses with recourse to feedback are judged to be of poorer quality than the baselines ones.

\subsection{With-Feedback Should Win} \label{sec:intersection}

While the previous section showed a discrepant trend between the JLM output and the ground truth, we note that preferring $r_{-f}$, even where feedback is helpful, is not necessarily erroneous. For example, there are cases were the improver manages to fix the corruption by itself, without feedback (\S\ref{sec:feedback_adrr}). For such cases, there is no reason to prefer the with-feedback variant. Moreover, it is possible that these without-feedback variants that solved the issue also improved other aspects of the response and should rightfully be preferred.

We therefore recompute the evaluation only on the subset of user queries where $r_{+f}$ successfully addressed the issue while $r_{-f}$ failed. Formally, this is defined as 
\begin{equation*}
\resizebox{\linewidth}{!}{%
$\begin{aligned}
\text{WFShouldWin}(M_i) &= \{q \mid J_{FA}(q,r_c,r_{+f},f)=\text{True}\} \\
&\quad \cap \{q \mid J_{FA}(q,r_c,r_{-f},f)=\text{False}\}
\end{aligned}$%
}
\end{equation*} 

On this subset, a judge is correct iff it prefers $r_{+f}$, as it is the only one that resolves the corruption.
Filtering for $q\in WFShouldWin$ for the various settings, we end up with $205$ and $574$ synthetic examples, and $215$ and $274$ naturalistic examples for $M_i^{large}$ and $M_i^{small}$ respectively.

The pairwise results highlights a frequent limitation in the automated evaluation of feedback signals: \textbf{even when filtering for examples where $r_{+f}$ should win, the judge  often fails in doing so.}

For the synthetic data, we find that the judge correctly prefers the with-feedback variant of the small improver in $80.9\%$ of the cases. For the large improver however, the judge correctly prefers the with-feedback variant in only $54.6\%$ of the cases, i.e there are $45.4\%$ cases where the judge makes an error and fails to prefer a valid response over an invalid one.
Similarly, in the natural data the judge correctly prefers the with-feedback variants in only $34\%$ of the cases for the large improver, and $54\%$ of the cases for the small improver.

\subsection{Models Struggle to Evaluate What They Cannot Improve} \label{sec:models_struggle}

Following the previous judge failure, we hypothesize that a model's limitations as an improver correlate with its limitations as an evaluator. Specifically, we suspect that a model will tend to fail as a judge on queries where it failed to self-improve.

To test this, we  compare the judge's accuracy on two sets of queries: those $M_i^{large}$ can self-improve, and those it cannot. The methodological challenge here is establishing a clear ground truth for the judgment task. If we were to use $M_i^{large}$'s own outputs for this pairwise comparison, the queries where it successfully resolved the corruption both with and without feedback would yield valid responses. Therefore, it would not be possible to determine on what cases the judge was correct.
Manually annotated ground truth is also infeasible (see \S\ref{sec:limitations}).

To resolve this, we decouple the query split from the evaluated responses by leveraging $M_i^{small}$. We restrict our evaluation entirely to the $WFShouldWin(M_i^{small})$ subset of the synthetic data. For these queries, we have a clear ground truth: the small improver's $r_{+f}$ successfully resolves the corruption, while its $r_{-f}$ fails. Therefore, in a pairwise comparison of these examples, $r_{+f}$ should always win.

We partition these ground-truth queries based on how the large improver performed on them, creating two subsets:

\begin{itemize}[leftmargin=1em, itemsep=0pt, topsep=0pt]
    \item \textbf{The self-improvable subset:} Queries where $M_i^{large}$ successfully resolved the corruption with or without feedback.
    \begin{equation*}
    \resizebox{\linewidth}{!}{%
    $\begin{aligned}
    \text{Both}(M_i^{large}) &= \{q \mid J_{FA}(q,r_c,r_{+f},f)=\text{True}\} \\
    &\quad \cap \{q \mid J_{FA}(q,r_c,r_{-f},f)=\text{True}\}
    \end{aligned}$%
    }
    \end{equation*}
    \item \textbf{The feedback-improvable subset:} Queries where $M_i^{large}$ could \textit{only} resolve the corruption when provided with feedback, namely $WFShouldWin(M_i^{large})$.
\end{itemize}

We then task $M_i^{large}$ to act as a judge, running a pairwise comparison between $r_{+f}$ and $r_{-f}$ generated by $M_i^{small}$ for both subsets. The $M_i^{large}$ judge correctly prefers $r_{+f}$ at high rates for both subsets: $87.7\%$ on the self-improvable subset, and $72.2\%$ on the feedbacl-improvable subset. However, its accuracy is $15.5$ percentage points lower on the feedback-dependent subset. This supports our hypothesis: \textbf{models exhibit correlated failure modes: an inability to independently improve a corrupted response corresponds to a decreased accuracy in evaluating it.}


\begin{figure}[t!]
\includegraphics[width=\linewidth]{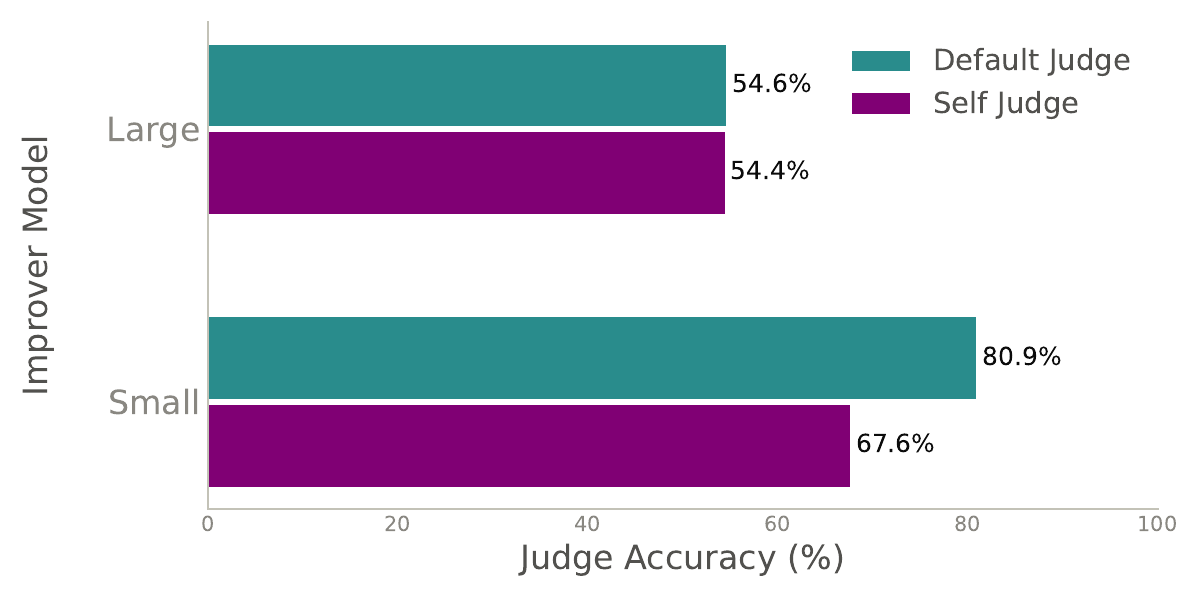}
\caption{Judge accuracy for the large and small improvers, with both default and self judges. While for the large improver there is no significant gap between the default and self judge, for the small improver with self judge we see for the first time that the judge struggle to prefer the correct response (the with-feedback variant).}
\label{fig:self_judge}
\end{figure}

\section{Further Analysis}
\subsection{Reproducing Correlated Failures in the Small Improver} \label{sec:self_judge}

To further investigate the relationship between a model's capabilities as an improver and as an evaluator, we assess the models in a self-judging setup where the improver also acts as the pairwise judge.

Figure~\ref{fig:self_judge} compares the evaluation accuracy of the default judge ({\it Gemini-3.1-Pro}) against the respective self-judges for the $WFShouldWin(M_i^{large})$ and $WFShouldWin(M_i^{small})$ subsets (\S\ref{sec:intersection}) of the synthetic data. 

This self-judging setup explains the different trends we saw for the large and small improvers. When using the default judge to evaluate the small model's subset, the evaluation appeared highly accurate (\S\ref{sec:intersection}). This occurred because of the substantial capability gap between the two models: issues that were hard for {\it Qwen3-8B} were still relatively easy for {\it Gemini-3.1-Pro}. Conversely, the capability gap between the default judge and the large improver is significantly narrower. Thus, the issues that {\it Gemini-3-Flash} failed to resolve were also difficult for {\it Gemini-3.1-Pro} to judge, making the shared limitations immediately visible during that comparison. Without the advantage of a more capable judge, misjudgment rates spike significantly, and we finally observe the same correlated failure trends previously seen with the large model.

The data confirms this dynamic. For the large improver, the default judge and the self-judge correctly prefer $r_{+f}$ at practically the same rate: $54.6\%$ and $54.4\%$ respectively.
However, when we use {\it Qwen3-8B} as a self-judge on its respective subset, a substantial performance gap emerges. While the default judge prefers the with-feedback variant in $80.8\%$ of the cases, the self-judge accuracy is only $67.6\%$ correct, $13$ points lower.

\begin{figure*}[t!]
    \centering
    \begin{subfigure}[b]{0.45\textwidth}
        \centering
        \includegraphics[width=\linewidth]{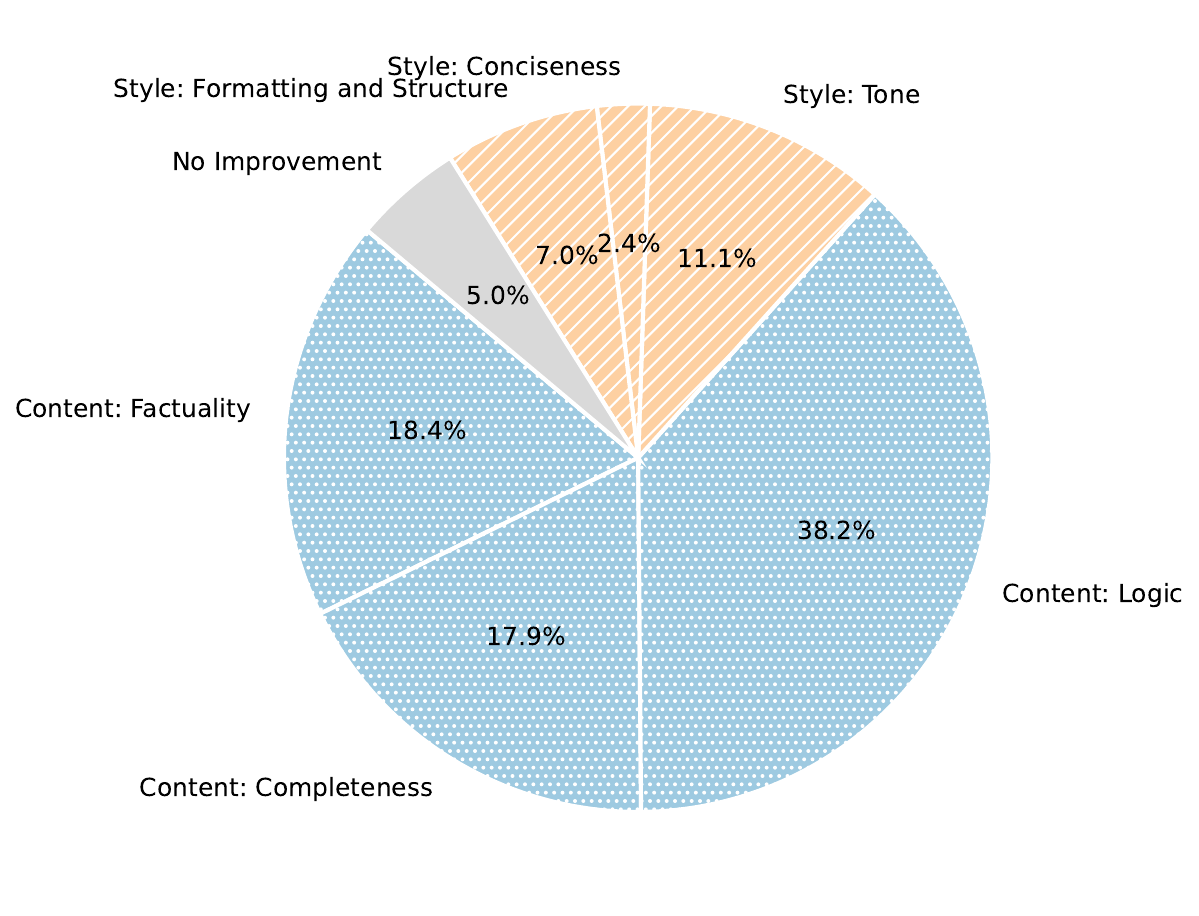}
        \caption{With Feedback}
        \label{fig:improve_with_feedback}
    \end{subfigure}
    \hfill
    \begin{subfigure}[b]{0.45\textwidth}
        \centering
        \includegraphics[width=\linewidth]{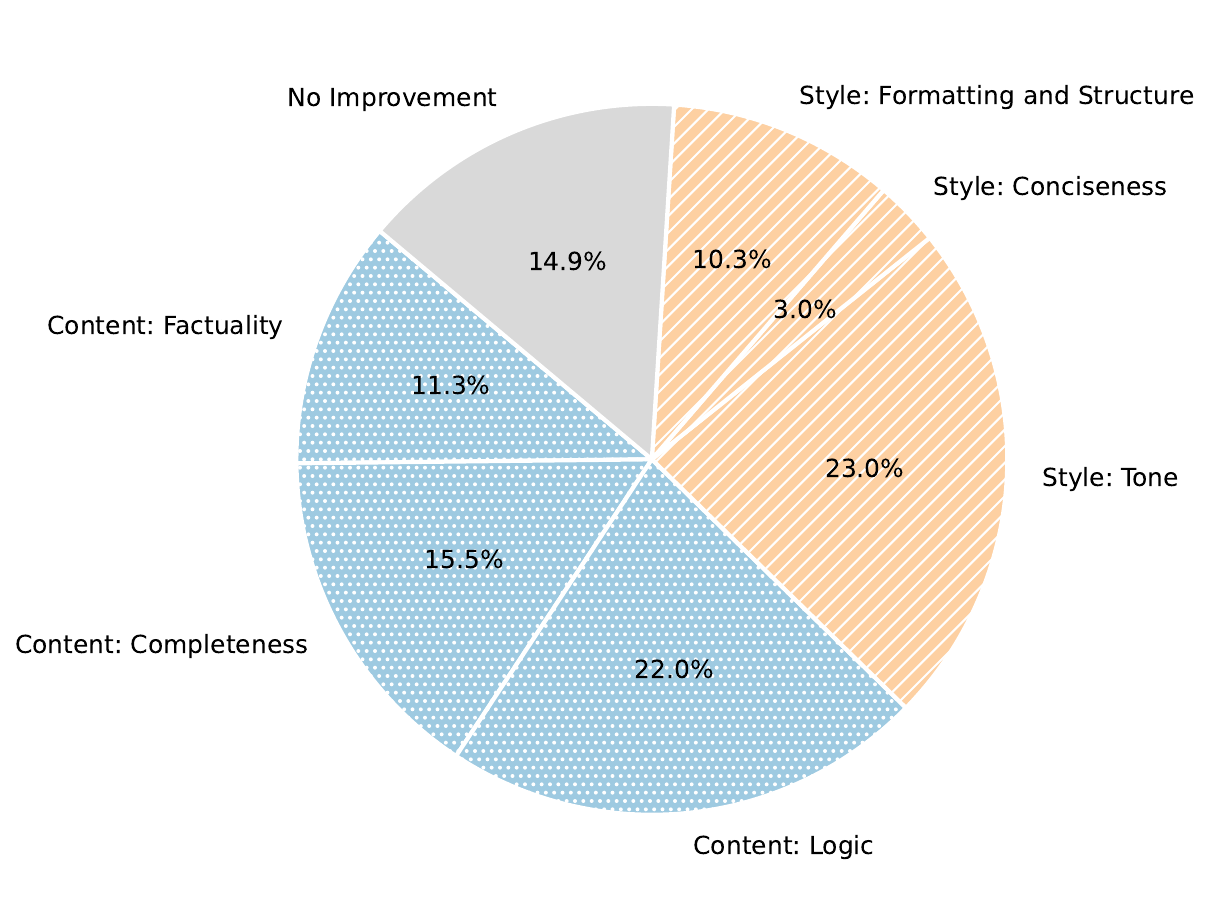}
        \caption{No Feedback}
        \label{fig:improve_no_feedback}
    \end{subfigure}

    \caption{Distribution of improvement types applied to corrupted model responses. Content-related improvements (factuality, completeness, and logic) are represented in shades of blue with a dotted texture. Style-related improvements (tone, conciseness, formatting) are represented in shades of orange with a diagonal line texture. Providing feedback significantly increases the proportion of substantive content improvements compared to stylistic edits.}
    \label{fig:improvement_types}
\end{figure*}

\subsection{Improvement Types}\label{sec:improvement_type}

How do improvements with and without feedback differ? This comparison may explain why the judge prefers still-corrupted responses.

We prompt a model to classify the $WFShouldWin(M_i^{large})$ attempted improvements of the full synthetic dataset (hard prompt + creative writing). The model categorizes each response into a \textit{No Improvement} option, or into one of six categories across two dimensions: \textit{Content} (Factuality, Completeness, Logic) and \textit{Style} (Tone, Conciseness, Formatting and Structure). See Appendix~\ref{app:improvement_type} for the full prompt.

Figure~\ref{fig:improvement_types} demonstrates that providing feedback increases the proportion of content improvements relative to style improvements. Additionally, the number of ``no improvement'' cases decreases when feedback is provided.

We conclude that \textbf{feedback yields higher-quality improvements focused on substantive content changes rather than stylistic edits. The pairwise judge, however, fails to recognize this and is biased towards stylistic changes.}

See Appendix~\ref{app:reimprove} for further evidence of judge bias towards stylistic edits. 


\begin{table}[htbp]
\centering
\small 
\setlength{\tabcolsep}{4pt} 
\begin{tabular}{lcccc}
\toprule
\textbf{Improver} & 
\begin{tabular}{@{}c@{}}\textbf{w/o.} \\ \textbf{Feedback}\end{tabular} & 
\begin{tabular}{@{}c@{}}\textbf{Full} \\ \textbf{Feedback}\end{tabular} & 
\begin{tabular}{@{}c@{}}\textbf{What} \\ \textbf{Problem}\end{tabular} & 
\begin{tabular}{@{}c@{}}\textbf{Problem} \\ \textbf{Exists}\end{tabular} \\
\midrule
\textbf{Large} & 86.3\% & 95.6\% & 92.8\% & 89.7\% \\
 & -- & \textit{$\Delta$+9.3\%} & \textit{$\Delta$+6.3\%} & \textit{$\Delta$+3.4\%} \\
\midrule
\textbf{Small} & 49.2\% & 80.8\% & 65.85\% & 59.5\% \\
 & -- & \textit{$\Delta$+31.6\%} & \textit{$\Delta$+16.7\%} & \textit{$\Delta$+10.3\%} \\
\bottomrule
\end{tabular}
\caption{Percentage of improved responses successfully addressing feedback, with different levels of feedback signal. Although the ``Full Feedback'' signal yields the largest delta ($9\%-32\%$), the weaker signals are still beneficial for the improvers.}
\label{tab:feedback_levels_comparison}
\end{table}

\subsection{Weaker Feedback Signal}

Our synthetic data experiments used feedback that tells the model how to fix the corruption ($f_{sol}$). Here we examine weaker feedback signals, that only point out the problem ($f_{prob}$), or just state that a problem exists ($f_{binary}$).

We generate $M_i(q,r_c,f_{prob})=r_{+f_{prob}}$ and $M_i(q,r_c,f_{binary})=r_{+f_{binary}}$ for $100$ queries for each corruption type ($400$ for each, $800$ total), and examine whether they better address the feedback over the without-feedback variant $r_{-f}$. Note that we provide $J_{IR}$ with $f_{sol}$ (and not $f_{prob}$ / $f_{binary}$ accordingly) as we want to provide it with sufficient information to verify whether the corruption was solved, thus we have $J_{FA}(q,r_c,f_{sol},r_{+f_{prob}})$ and $J_{FA}(q,r_c,f_{sol},r_{+f_{binary}})$.

Table~\ref{tab:feedback_levels_comparison} demonstrates that both the feedback that points to the problem (``What Problem''), and the feedback that only states that there is a problem (``Problem exists'') solve the corruption in a higher rate than the no-feedback variant. However, as expected, their margin is smaller. While for the full feedback (\S\ref{sec:feedback_adrr}) the improvers resolved about $9\%-32\%$ more cases than without any feedback, ``What Problem'' feedback improves about $6\%-17\%$ more cases, and ``Problem exists'' improves only about $3\%-10\%$ more cases. 
Also, similar to the full feedback, the small improver benefits more from the feedback.  

We conclude that pointing to a problem, or just indicating that a problem exists are valuable, even if less than the full feedback.

\section{Related Work}

A related line of work emphasized the utility of \emph{natural language feedback} \citep{hancock-etal-2019-learning, yan-etal-2023-learning, enhancing_performance, sreedhar-etal-2020-learning}. In such works, the user is asked to provide free-text feedback (unlike the more standard binary feedback or other close forms).
Later works \citep{naturallyoccurringfeedback, lin-etal-2024-interpretable, shi-etal-2026-wildfeedback} take a step further and look for feedback that spontaneously occurs in the conversations, without explicitly asking the user to provide it, and use it for model training \citep{buening2026aligning, jin2025era}.

\citet{liu-etal-2025-user} studied the question of whether naturally occurring user feedback 
can indeed improve model performance. They used a large model to create ``regeneration with semantics'' responses that used human feedback, and ``regeneration from scratch'' responses as a baseline, similar to our with-feedback vs. without-feedback responses. Using a reward model to compare them against each other, they found that the regenerations from scratch outperformed the semantics ones. This is in line with our results, that for the large improver the without-feedback responses outperformed the with-feedback responses (\S\ref{sec:results_pairwise}). However, our analysis revealed that this is due to a problem in the evaluation, not the signal's usefulness (\S\ref{sec:intersection}).

In our work we demonstrated how the well used "LLM as a Judge" evaluation \citep{chiang-lee-2023-large, liu-etal-2023-g, zhu2023judgelm} fails to prefer the valid response over the corrupted one. Previous works already show that this method suffers from biases, such as position, verbosity and self-preference \citep{wang-etal-2024, saito2023verbosity, koo-etal-2024-benchmarking, don-yehiya-etal-2026-mediocrity}.

In the scope of this work we improve the model in test-time \citep{madaan2023self}, without further training. Another line of work investigates self-improve the model via training \citep{NEURIPS2025_6b41e04c, huang-etal-2023-large, 3692070.3694459}.

\section{Discussion}
Through validation on both synthetic and naturalistic data, we established that while LLMs can resolve numerous issues autonomously, incorporating user feedback enables them to resolve $9\%-32\%$ more cases compared to baselines without feedback.
However, this improvement is not captured during evaluation:  despite a higher resolution rate when feedback is applied, pairwise evaluations frequently favor the without-feedback variants. This explains the prevailing misconception that feedback does not provide a useful signal.

Preference for the without-feedback variant may be warranted in instances where both variants successfully resolve the error, as the without-feedback variant may introduce other general improvements (\S\ref{sec:improvement_type}). However, in cases where only $r_{+f}$ variant resolves the corruption, a preference for $r_{-f}$ or a tie, is a clear error on the LLM judge's behalf.

Further analysis reveals a circular pattern: the judge frequently fails to correctly judge cases that the LLM improver failed to improve on its own. This yields a bias against the feedback signal.

We speculate that this systematic bias may be, oddly enough, evidence for the utility of naturally occurring feedback. The fact that LLM judges fail to recognize feedback-informed improvements might suggest that the feedback provides a unique signal that is not encoded in the model's weights. Thus, it is reasonable to assume that it can not be acquired through standard techniques such as model distillation or iterative self-improvement.

Although demonstrated in an inference-only setting, we expect our findings to persist in training as well.
Consequently, future work should prioritize the development of more robust evaluation frameworks that accurately reflect the utility of feedback. We urge the community to look beyond standard LLM judgment paradigms and explore specialized reward models, interactive evaluation metrics, or targeted alignment strategies designed specifically to break this circular bias. Addressing this fundamental evaluation gap is critical; without reliable mechanisms to appropriately credit feedback-driven resolutions, we risk artificially impeding the development of open models, developed with open user data. 

\section*{Limitations}\label{sec:limitations}
Our manual annotation revealed that the corruption process is less effective for the creative writing category of our synthetic data.

Therefore, to ensure reliability, the main text focuses exclusively on the math and code domains. However, results for the full synthetic dataset, including creative writing, are available in the appendix (\S\ref{app:full_results}). The overall trends persist, which, along with our multi-domain naturalistic data, indicate that our findings generalize beyond math and code domains.

To ensure reliability, we manually annotated representative samples from all automated stages of our pipeline: the corruption process, feedback relevancy filtering, and feedback addressing evaluation. 
However, we did not manually annotate the pairwise evaluation setting, as that proved to be prohibitively difficult for humans. We did, however, conduct a small-scale human annotation task which confirmed that manual pairwise evaluation is highly challenging. The sheer length of the responses makes it difficult to track minute details, and the diverse topics necessitate broad domain expertise.

\section*{Acknowledgments}
This research was partly supported by a grant from
the Israel Science Foundation (grant no. 2912/25).
Special thanks to our annotators, Nicole Gruber and Yoav Schmidt.

\bibliography{custom}

@inproceedings{liu-etal-2025-user,
    title = "User Feedback in Human-{LLM} Dialogues: A Lens to Understand Users But Noisy as a Learning Signal",
    author = "Liu, Yuhan  and
      Zhang, Michael JQ  and
      Choi, Eunsol",
    editor = "Christodoulopoulos, Christos  and
      Chakraborty, Tanmoy  and
      Rose, Carolyn  and
      Peng, Violet",
    booktitle = "Proceedings of the 2025 Conference on Empirical Methods in Natural Language Processing",
    month = nov,
    year = "2025",
    address = "Suzhou, China",
    publisher = "Association for Computational Linguistics",
    url = "https://aclanthology.org/2025.emnlp-main.133/",
    doi = "10.18653/v1/2025.emnlp-main.133",
    pages = "2666--2681",
    ISBN = "979-8-89176-332-6"
}

@inproceedings{don-yehiya-sharelm,
    title = "The {S}hare{LM} Collection and Plugin: Contributing Human-Model Chats for the Benefit of the Community",
    author = "Don-Yehiya, Shachar  and
      Choshen, Leshem  and
      Abend, Omri",
    editor = "Mishra, Pushkar  and
      Muresan, Smaranda  and
      Yu, Tao",
    booktitle = "Proceedings of the 63rd Annual Meeting of the Association for Computational Linguistics (Volume 3: System Demonstrations)",
    month = jul,
    year = "2025",
    address = "Vienna, Austria",
    publisher = "Association for Computational Linguistics",
    url = "https://aclanthology.org/2025.acl-demo.17/",
    pages = "167--177",
    ISBN = "979-8-89176-253-4"
}

@misc{naturallyoccurringfeedback,
      title={Naturally Occurring Feedback is Common, Extractable and Useful}, 
      author={Shachar Don-Yehiya and Leshem Choshen and Omri Abend},
      year={2025},
      eprint={2407.10944},
      archivePrefix={arXiv},
      primaryClass={cs.CL},
      url={https://arxiv.org/abs/2407.10944}, 
}

@article{chiang2024chatbot,
  title={Chatbot arena: An open platform for evaluating llms by human preference},
  author={Chiang, Wei-Lin and Zheng, Lianmin and Sheng, Ying and Angelopoulos, Anastasios Nikolas and Li, Tianle and Li, Dacheng and Zhang, Hao and Zhu, Banghua and Jordan, Michael and Gonzalez, Joseph E and others},
  journal={arXiv preprint arXiv:2403.04132},
  year={2024}
}

@article{li2024crowdsourced,
  title={From Crowdsourced Data to High-Quality Benchmarks: Arena-Hard and BenchBuilder Pipeline},
  author={Li, Tianle and Chiang, Wei-Lin and Frick, Evan and Dunlap, Lisa and Wu, Tianhao and Zhu, Banghua and Gonzalez, Joseph E and Stoica, Ion},
  journal={arXiv preprint arXiv:2406.11939},
  year={2024}
}

@inproceedings{lin-etal-2024-interpretable,
    title = "Interpretable User Satisfaction Estimation for Conversational Systems with Large Language Models",
    author = "Lin, Ying-Chun  and
      Neville, Jennifer  and
      Stokes, Jack  and
      Yang, Longqi  and
      Safavi, Tara  and
      Wan, Mengting  and
      Counts, Scott  and
      Suri, Siddharth  and
      Andersen, Reid  and
      Xu, Xiaofeng  and
      Gupta, Deepak  and
      Jauhar, Sujay Kumar  and
      Song, Xia  and
      Buscher, Georg  and
      Tiwary, Saurabh  and
      Hecht, Brent  and
      Teevan, Jaime",
    editor = "Ku, Lun-Wei  and
      Martins, Andre  and
      Srikumar, Vivek",
    booktitle = "Proceedings of the 62nd Annual Meeting of the Association for Computational Linguistics (Volume 1: Long Papers)",
    month = aug,
    year = "2024",
    address = "Bangkok, Thailand",
    publisher = "Association for Computational Linguistics",
    url = "https://aclanthology.org/2024.acl-long.598/",
    doi = "10.18653/v1/2024.acl-long.598",
    pages = "11100--11115"
}

@inproceedings{shi-etal-2026-wildfeedback,
    title = "{W}ild{F}eedback: Aligning {LLM}s With In-situ User Interactions And Feedback",
    author = "Shi, Taiwei  and
      Wang, Zhuoer  and
      Yang, Longqi  and
      Lin, Ying-Chun  and
      He, Zexue  and
      Wan, Mengting  and
      Zhou, Pei  and
      Jauhar, Sujay Kumar  and
      Chen, Sihao  and
      Xia, Shan  and
      Zhang, Hongfei  and
      Zhao, Jieyu  and
      Xu, Xiaofeng  and
      Song, Xia  and
      Neville, Jennifer",
    editor = "Liakata, Maria  and
      Moreira, Viviane P.  and
      Zhang, Jiajun  and
      Jurgens, David",
    booktitle = "Proceedings of the 64th Annual Meeting of the {A}ssociation for {C}omputational {L}inguistics (Volume 1: Long Papers)",
    month = jul,
    year = "2026",
    address = "San Diego, California, United States",
    publisher = "Association for Computational Linguistics",
    url = "https://aclanthology.org/2026.acl-long.1701/",
    doi = "10.18653/v1/2026.acl-long.1701",
    pages = "36701--36725",
    ISBN = "979-8-89176-390-6"
}

@article{buening2026aligning,
  title={Aligning language models from user interactions},
  author={Buening, Thomas Kleine and H{\"u}botter, Jonas and P{\'a}sztor, Barna and Shenfeld, Idan and Ramponi, Giorgia and Krause, Andreas},
  journal={arXiv preprint arXiv:2603.12273},
  year={2026}
}

@article{jin2025era,
  title={The era of real-world human interaction: Rl from user conversations},
  author={Jin, Chuanyang and Xu, Jing and Liu, Bo and Tao, Leitian and Golovneva, Olga and Shu, Tianmin and Zhao, Wenting and Li, Xian and Weston, Jason},
  journal={arXiv preprint arXiv:2509.25137},
  year={2025}
}

@article{DonYehiya_2025a, title={The future of open human feedback}, volume={7}, DOI={10.1038/s42256-025-01038-2}, number={6}, journal={Nature Machine Intelligence}, author={Don-Yehiya, Shachar and Burtenshaw, Ben and Fernandez Astudillo, Ramon and Osborne, Cailean and Jaiswal, Mimansa and Kuo, Tzu-Sheng and Zhao, Wenting and Shenfeld, Idan and Peng, Andi and Yurochkin, Mikhail and et al.}, year={2025}, month={Jun}, pages={825–835}}

@inproceedings{hancock-etal-2019-learning,
    title = "Learning from Dialogue after Deployment: Feed Yourself, Chatbot!",
    author = "Hancock, Braden  and
      Bordes, Antoine  and
      Mazare, Pierre-Emmanuel  and
      Weston, Jason",
    editor = "Korhonen, Anna  and
      Traum, David  and
      M{\`a}rquez, Llu{\'\i}s",
    booktitle = "Proceedings of the 57th Annual Meeting of the Association for Computational Linguistics",
    month = jul,
    year = "2019",
    address = "Florence, Italy",
    publisher = "Association for Computational Linguistics",
    url = "https://aclanthology.org/P19-1358",
    doi = "10.18653/v1/P19-1358",
    pages = "3667--3684",
}

@inproceedings{yan-etal-2023-learning,
    title = "Learning to Simulate Natural Language Feedback for Interactive Semantic Parsing",
    author = "Yan, Hao  and
      Srivastava, Saurabh  and
      Tai, Yintao  and
      Wang, Sida I.  and
      Yih, Wen-tau  and
      Yao, Ziyu",
    editor = "Rogers, Anna  and
      Boyd-Graber, Jordan  and
      Okazaki, Naoaki",
    booktitle = "Proceedings of the 61st Annual Meeting of the Association for Computational Linguistics (Volume 1: Long Papers)",
    month = jul,
    year = "2023",
    address = "Toronto, Canada",
    publisher = "Association for Computational Linguistics",
    url = "https://aclanthology.org/2023.acl-long.177",
    doi = "10.18653/v1/2023.acl-long.177",
    pages = "3149--3170",
}

@inproceedings{enhancing_performance,
author = {Jayalath, Hemadri and Ramaswamy, Lakshmish},
title = {Enhancing Performance of Operationalized Machine Learning Models by Analyzing User Feedback},
year = {2022},
isbn = {9781450387415},
publisher = {Association for Computing Machinery},
address = {New York, NY, USA},
url = {https://doi.org/10.1145/3531232.3531261},
doi = {10.1145/3531232.3531261},
booktitle = {Proceedings of the 2022 4th International Conference on Image, Video and Signal Processing},
pages = {197–203},
numpages = {7},
location = {Singapore, Singapore},
series = {IVSP '22}
}

@inproceedings{sreedhar-etal-2020-learning,
    title = "Learning Improvised Chatbots from Adversarial Modifications of Natural Language Feedback",
    author = "Sreedhar, Makesh Narsimhan  and
      Ni, Kun  and
      Reddy, Siva",
    editor = "Cohn, Trevor  and
      He, Yulan  and
      Liu, Yang",
    booktitle = "Findings of the Association for Computational Linguistics: EMNLP 2020",
    month = nov,
    year = "2020",
    address = "Online",
    publisher = "Association for Computational Linguistics",
    url = "https://aclanthology.org/2020.findings-emnlp.221",
    doi = "10.18653/v1/2020.findings-emnlp.221",
    pages = "2445--2453",
}

@article{zhu2023judgelm,
  title={Judgelm: Fine-tuned large language models are scalable judges},
  author={Zhu, Lianghui and Wang, Xinggang and Wang, Xinlong},
  journal={arXiv preprint arXiv:2310.17631},
  year={2023}
}

@inproceedings{don-yehiya-etal-2026-mediocrity,
    title = "Mediocrity is the key for {LLM} as a Judge Anchor Selection",
    author = "Don-Yehiya, Shachar  and
      Yehudai, Asaf  and
      Choshen, Leshem  and
      Abend, Omri",
    editor = "Liakata, Maria  and
      Moreira, Viviane P.  and
      Zhang, Jiajun  and
      Jurgens, David",
    booktitle = "Proceedings of the 64th Annual Meeting of the {A}ssociation for {C}omputational {L}inguistics (Volume 1: Long Papers)",
    month = jul,
    year = "2026",
    address = "San Diego, California, United States",
    publisher = "Association for Computational Linguistics",
    url = "https://aclanthology.org/2026.acl-long.706/",
    doi = "10.18653/v1/2026.acl-long.706",
    pages = "15491--15513",
    ISBN = "979-8-89176-390-6"
}

@article{saito2023verbosity,
  title={Verbosity bias in preference labeling by large language models},
  author={Saito, Keita and Wachi, Akifumi and Wataoka, Koki and Akimoto, Youhei},
  journal={arXiv preprint arXiv:2310.10076},
  year={2023}
}

@inproceedings{wang-etal-2024,
    title = "Large Language Models are not Fair Evaluators",
    author = "Wang, Peiyi  and
      Li, Lei  and
      Chen, Liang  and
      Cai, Zefan  and
      Zhu, Dawei  and
      Lin, Binghuai  and
      Cao, Yunbo  and
      Kong, Lingpeng  and
      Liu, Qi  and
      Liu, Tianyu  and
      Sui, Zhifang",
    editor = "Ku, Lun-Wei  and
      Martins, Andre  and
      Srikumar, Vivek",
    booktitle = "Proceedings of the 62nd Annual Meeting of the Association for Computational Linguistics (Volume 1: Long Papers)",
    month = aug,
    year = "2024",
    address = "Bangkok, Thailand",
    publisher = "Association for Computational Linguistics",
    url = "https://aclanthology.org/2024.acl-long.511/",
    doi = "10.18653/v1/2024.acl-long.511",
    pages = "9440--9450"
}

@inproceedings{koo-etal-2024-benchmarking,
    title = "Benchmarking Cognitive Biases in Large Language Models as Evaluators",
    author = "Koo, Ryan  and
      Lee, Minhwa  and
      Raheja, Vipul  and
      Park, Jong Inn  and
      Kim, Zae Myung  and
      Kang, Dongyeop",
    editor = "Ku, Lun-Wei  and
      Martins, Andre  and
      Srikumar, Vivek",
    booktitle = "Findings of the Association for Computational Linguistics: ACL 2024",
    month = aug,
    year = "2024",
    address = "Bangkok, Thailand",
    publisher = "Association for Computational Linguistics",
    url = "https://aclanthology.org/2024.findings-acl.29/",
    doi = "10.18653/v1/2024.findings-acl.29",
    pages = "517--545"
}

@inproceedings{chiang-lee-2023-large,
    title = "Can Large Language Models Be an Alternative to Human Evaluations?",
    author = "Chiang, Cheng-Han  and
      Lee, Hung-yi",
    editor = "Rogers, Anna  and
      Boyd-Graber, Jordan  and
      Okazaki, Naoaki",
    booktitle = "Proceedings of the 61st Annual Meeting of the Association for Computational Linguistics (Volume 1: Long Papers)",
    month = jul,
    year = "2023",
    address = "Toronto, Canada",
    publisher = "Association for Computational Linguistics",
    url = "https://aclanthology.org/2023.acl-long.870/",
    doi = "10.18653/v1/2023.acl-long.870",
    pages = "15607--15631"
}

@inproceedings{liu-etal-2023-g,
    title = "{G}-Eval: {NLG} Evaluation using Gpt-4 with Better Human Alignment",
    author = "Liu, Yang  and
      Iter, Dan  and
      Xu, Yichong  and
      Wang, Shuohang  and
      Xu, Ruochen  and
      Zhu, Chenguang",
    editor = "Bouamor, Houda  and
      Pino, Juan  and
      Bali, Kalika",
    booktitle = "Proceedings of the 2023 Conference on Empirical Methods in Natural Language Processing",
    month = dec,
    year = "2023",
    address = "Singapore",
    publisher = "Association for Computational Linguistics",
    url = "https://aclanthology.org/2023.emnlp-main.153/",
    doi = "10.18653/v1/2023.emnlp-main.153",
    pages = "2511--2522"
}

@article{madaan2023self,
  title={Self-refine: Iterative refinement with self-feedback},
  author={Madaan, Aman and Tandon, Niket and Gupta, Prakhar and Hallinan, Skyler and Gao, Luyu and Wiegreffe, Sarah and Alon, Uri and Dziri, Nouha and Prabhumoye, Shrimai and Yang, Yiming and others},
  journal={Advances in neural information processing systems},
  volume={36},
  pages={46534--46594},
  year={2023}
}

@inproceedings{NEURIPS2025_6b41e04c,
 author = {Zweiger, Adam and Pari, Jyo and Guo, Han and Kim, Yoon and Agrawal, Pulkit},
 booktitle = {Advances in Neural Information Processing Systems},
 editor = {D. Belgrave and C. Zhang and H. Lin and R. Pascanu and P. Koniusz and M. Ghassemi and N. Chen},
 pages = {74084--74115},
 publisher = {Curran Associates, Inc.},
 title = {Self-Adapting Language Models},
 url = {https://proceedings.neurips.cc/paper_files/paper/2025/file/6b41e04c41726e2a60e456d0a2b961ab-Paper-Conference.pdf},
 volume = {38},
 year = {2025}
}

@inproceedings{huang-etal-2023-large,
    title = "Large Language Models Can Self-Improve",
    author = "Huang, Jiaxin  and
      Gu, Shixiang  and
      Hou, Le  and
      Wu, Yuexin  and
      Wang, Xuezhi  and
      Yu, Hongkun  and
      Han, Jiawei",
    editor = "Bouamor, Houda  and
      Pino, Juan  and
      Bali, Kalika",
    booktitle = "Proceedings of the 2023 Conference on Empirical Methods in Natural Language Processing",
    month = dec,
    year = "2023",
    address = "Singapore",
    publisher = "Association for Computational Linguistics",
    url = "https://aclanthology.org/2023.emnlp-main.67/",
    doi = "10.18653/v1/2023.emnlp-main.67",
    pages = "1051--1068"
}

@inproceedings{3692070.3694459,
author = {Yuan, Weizhe and Pang, Richard Yuanzhe and Cho, Kyunghyun and Li, Xian and Sukhbaatar, Sainbayar and Xu, Jing and Weston, Jason},
title = {Self-rewarding language models},
year = {2024},
publisher = {JMLR.org},
booktitle = {Proceedings of the 41st International Conference on Machine Learning},
articleno = {2389},
numpages = {19},
location = {Vienna, Austria},
series = {ICML'24}
}

@article{team2023gemini,
  title={Gemini: a family of highly capable multimodal models},
  author={Team, Gemini and Anil, Rohan and Borgeaud, Sebastian and Alayrac, Jean-Baptiste and Yu, Jiahui and Soricut, Radu and Schalkwyk, Johan and Dai, Andrew M and Hauth, Anja and Millican, Katie and others},
  journal={arXiv preprint arXiv:2312.11805},
  year={2023}
}

@article{agarwal2025gpt,
  title={gpt-oss-120b \& gpt-oss-20b model card},
  author={Agarwal, Sandhini and Ahmad, Lama and Ai, Jason and Altman, Sam and Applebaum, Andy and Arbus, Edwin and Arora, Rahul K and Bai, Yu and Baker, Bowen and Bao, Haiming and others},
  journal={arXiv preprint arXiv:2508.10925},
  year={2025}
}

@article{yang2025qwen3,
  title={Qwen3 technical report},
  author={Yang, An and Li, Anfeng and Yang, Baosong and Zhang, Beichen and Hui, Binyuan and Zheng, Bo and Yu, Bowen and Gao, Chang and Huang, Chengen and Lv, Chenxu and others},
  journal={arXiv preprint arXiv:2505.09388},
  year={2025}
}

@book{pickering_garrod_2021, place={Cambridge}, title={Understanding Dialogue: Language Use and Social Interaction}, DOI={10.1017/9781108610728}, publisher={Cambridge University Press}, author={Pickering, Martin J. and Garrod, Simon}, year={2021}}

@article{VRANJES201815,
title = {Dual feedback in interpreter-mediated interactions: On the role of gaze in the production of listener responses},
journal = {Journal of Pragmatics},
volume = {134},
pages = {15-30},
year = {2018},
issn = {0378-2166},
doi = {https://doi.org/10.1016/j.pragma.2018.06.002},
url = {https://www.sciencedirect.com/science/article/pii/S0378216617303879},
author = {Jelena Vranjes and Geert Brône and Kurt Feyaerts}
}

@article{doi:10.1080/10904018.2010.508675,
author = {Janet Beavin Bavelas and Jennifer Gerwing},
title = {The Listener as Addressee in Face-to-Face Dialogue},
journal = {International Journal of Listening},
volume = {25},
number = {3},
pages = {178--198},
year = {2011},
publisher = {Routledge},
doi = {10.1080/10904018.2010.508675},


URL = { 
    
        https://doi.org/10.1080/10904018.2010.508675
    
    

},
eprint = { 
    
        https://doi.org/10.1080/10904018.2010.508675
    
    

}

}

@article{bassiri2011interactional,
  title={Interactional feedback and the impact of attitude and motivation on noticing l2 form},
  author={Bassiri, Mohammad Amin},
  journal={English Language and Literature Studies},
  volume={1},
  number={2},
  pages={61},
  year={2011},
  publisher={Canadian Center of Science and Education}
}

@article{5a61a38b-79fc-30d4-a76e-8a66a7707021,
 ISSN = {10530819, 15733513},
 URL = {http://www.jstor.org/stable/41826861},
 author = {Margaret G. Werts and Mark Wolery and Ariane Holcombe and David L. Gast},
 journal = {Journal of Behavioral Education},
 number = {1},
 pages = {55--75},
 publisher = {Springer},
 title = {Instructive Feedback: Review of Parameters and Effects},
 urldate = {2024-05-24},
 volume = {5},
 year = {1995}
}

\appendix

\section{Full Results} \label{app:full_results}
We provide here the full results, including the creative writing category.

We use the judge model $J_{IR}$ (\S\ref{sec:eval_pairwise}) to compare the original response $r$ and the corrupted one $r_c$. Table~\ref{tab:original_vs_corrupted_hard_prompt} confirms that in this clean setup, with minimal changes introduced by the corruption model, the judge correctly prefers the original response.

Table \ref{tab:feedback_addressing_hard_prompt} presents the issue resolution rates for the hard prompt category (synthetic data).
Table \ref{tab:feedback_addressing_full_synthetic_data} present the issue resolution rates for the full synthetic data, i.e., hard prompt + creative writing. We see that the trends remain the same, with a consistent positive delta between the without-feedback and with-feedback variants.

Table \ref{tab:original_vs_corrupted_hard_prompt} presents the judge pairwise preferences results for the hard prompt category of the synthetic data.
Table \ref{tab:with_feedback_vs_whithout_feedback_full} presents the pairwise results for the full synthetic data. Also here, the trends remain the same, with even larger effect (the preference of the without-feedback variant is stronger).

Table \ref{tab:judge_comparison_real_user_overall} presents the judge pairwise preferences results for the real user data. Table \ref{tab:judge_comparison_real_user_subset} presents the results for the \textit{WFShouldWin} subset of the user data.

\begin{figure*}[t!]
    \centering
    \begin{subfigure}[b]{0.45\textwidth}
        \centering
        \includegraphics[width=\linewidth]{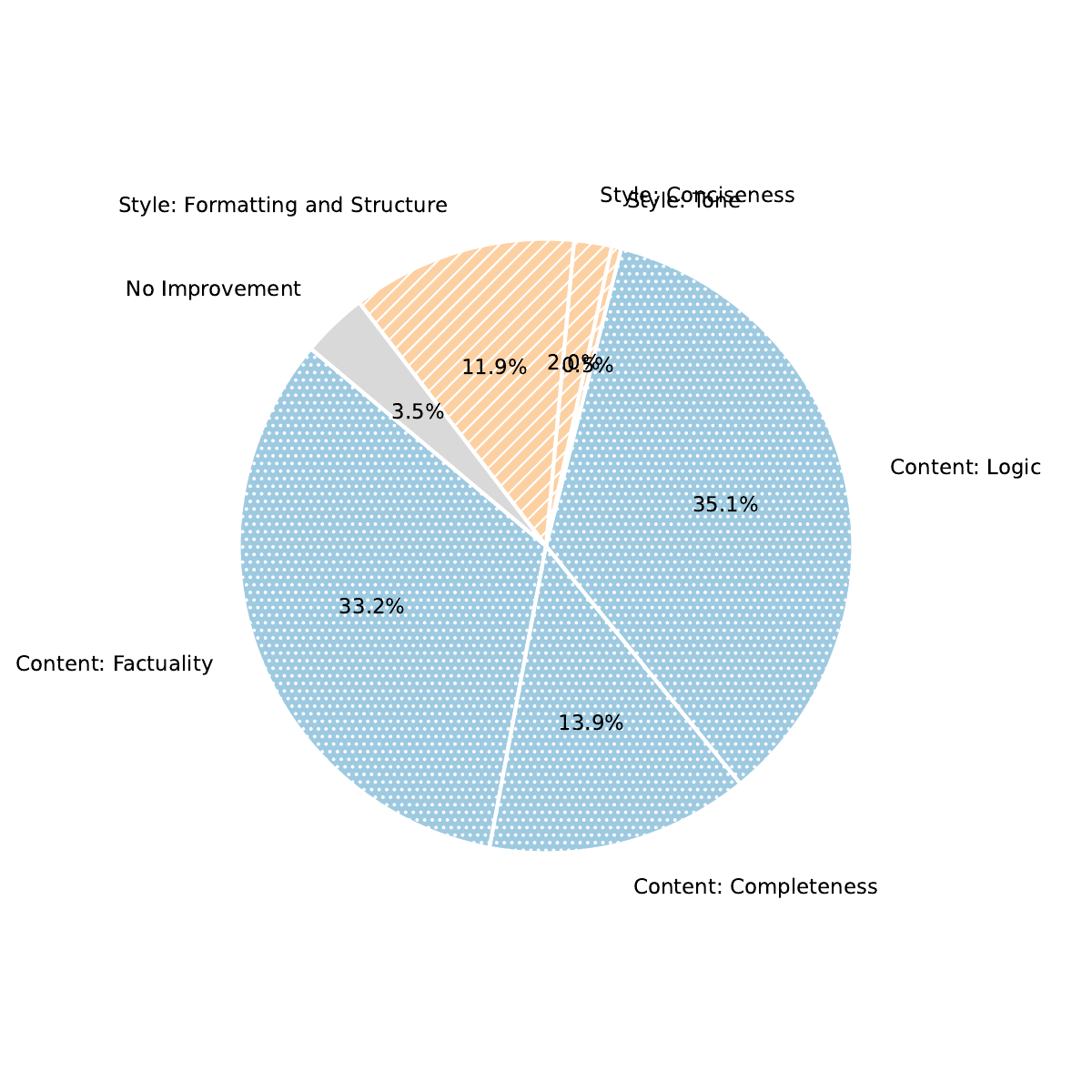}
        \caption{With Feedback}
        \label{fig:improve_with_feedback_hard_prompt}
    \end{subfigure}
    \hfill
    \begin{subfigure}[b]{0.45\textwidth}
        \centering
        \includegraphics[width=\linewidth]{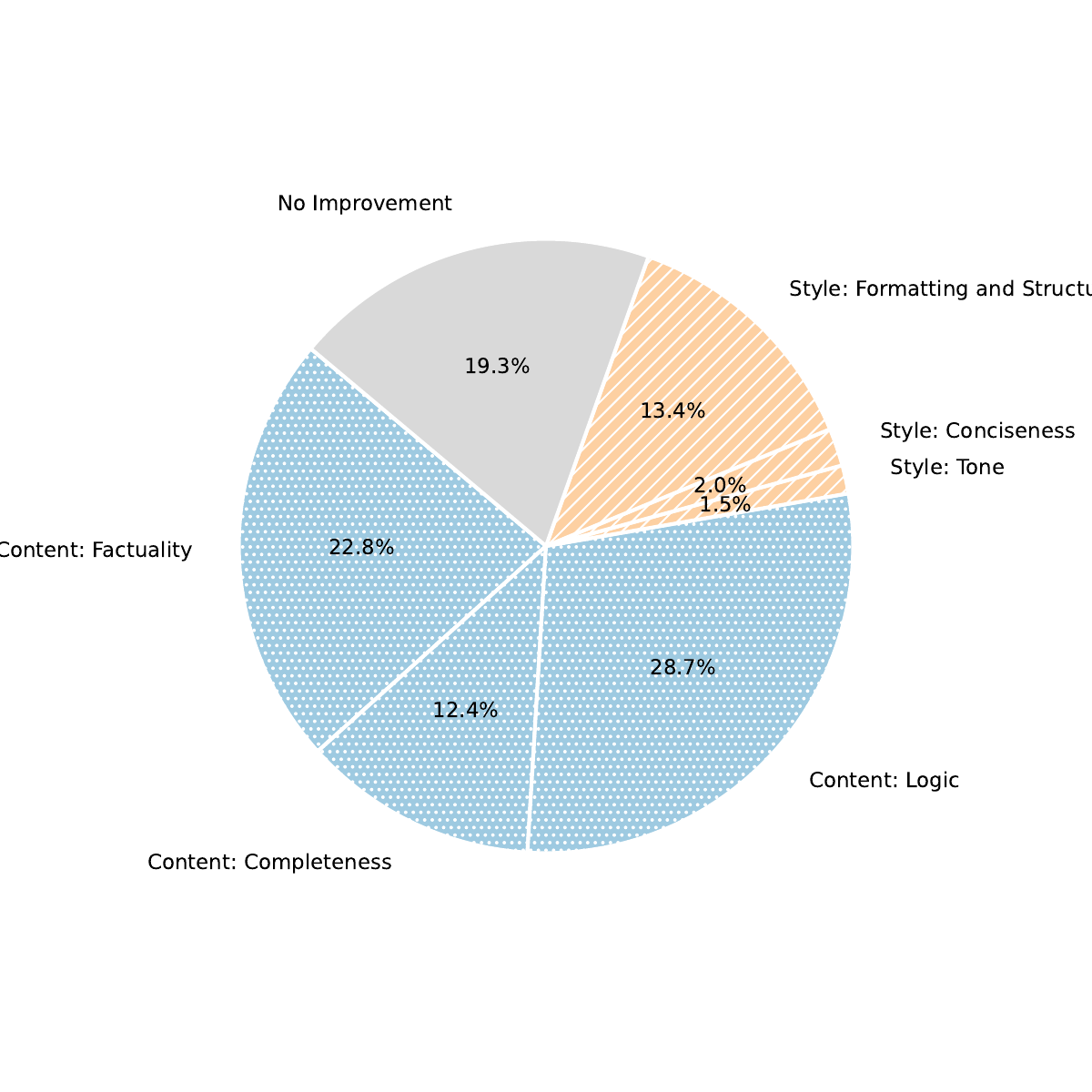}
        \caption{No Feedback}
        \label{fig:improve_no_feedback_hard_prompt}
    \end{subfigure}

    \caption{Distribution of improvement types applied to corrupted model responses for the hard prompt category. Content-related improvements (factuality, completeness, and logic) are represented in shades of blue with a dotted texture. Style-related improvements (tone, conciseness, formatting) are represented in shades of orange with a diagonal line texture. Providing feedback significantly increases the proportion of substantive content improvements compared to stylistic edits.}
    \label{fig:improvement_types_hard_prompt}
\end{figure*}

Fig.~\ref{fig:improvement_types_hard_prompt} presents the improvement type results for the hard prompt category. 

\begin{table}[t!]
\centering
\resizebox{\columnwidth}{!}{%
\begin{tabular}{lcccc}
\toprule
& \textbf{Causality} & \textbf{Crucial} & \textbf{Entity} & \textbf{Operator} \\
& \textbf{Inversion} & \textbf{Omission} & \textbf{Swap} & \textbf{Reversal} \\
\midrule
\textbf{Original} & 96.6\% & 98.7\% & 99.4\% & 99.2\% \\
\textbf{Corrupted} & 0.9\% & 1.1\% & 0.4\% & 0.8\% \\
\textbf{Tie} & 2.6\% & 0.2\% & 0.2\% & 0.0\% \\
\bottomrule
\end{tabular}%
}
\caption{Judge Comparison: Original vs. Corrupted Responses}
\label{tab:original_vs_corrupted_hard_prompt}
\end{table}

\begin{table*}[htbp]
\centering
\small 
\setlength{\tabcolsep}{4pt} 
\begin{tabular}{llccccc}
\toprule
\textbf{Improver} & \textbf{Variant} & \textbf{Causality} & \textbf{Crucial} & \textbf{Entity} & \textbf{Operator} & \textbf{Average} \\
& & \textbf{Inversion} & \textbf{Omission} & \textbf{Swap} & \textbf{Reversal} & \\
\midrule
\multirow{3}{*}{\textbf{Large}} 
& w/o feedback & 86.6\% & 85.0\% & 85.1\% & 88.5\% & 86.3\% \\
& w/ feedback & 97.2\% & 94.9\% & 94.8\% & 95.5\% & 95.6\% \\
&  & \textit{$\Delta$+10.6\%} & \textit{$\Delta$+9.9\%} & \textit{$\Delta$+9.7\%} & \textit{$\Delta$+7.0\%} & \textit{$\Delta$+9.3\%} \\
\midrule
\multirow{3}{*}{\textbf{Small}} 
& w/o feedback & 52.6\% & 47.5\% & 47.4\% & 49.4\% & 49.2\% \\
& w/ feedback & 78.9\% & 79.1\% & 83.1\% & 82.1\% & 80.8\% \\
&  & \textit{$\Delta$+26.3\%} & \textit{$\Delta$+31.6\%} & \textit{$\Delta$+35.7\%} & \textit{$\Delta$+32.7\%} & \textit{$\Delta$+31.6\%} \\
\bottomrule
\end{tabular}
\caption{Percentage of improved responses successfully addressing issues for the hard prompt category of the synthetic data. We observe a consistent positive delta between the without-feedback and with-feedback variants across the four corruption types.
}
\label{tab:feedback_addressing_hard_prompt}
\end{table*}

\begin{table*}[htbp]
\centering
\small 
\setlength{\tabcolsep}{4pt} 
\begin{tabular}{llccccc}
\toprule
\textbf{Improver} & \textbf{Variant} & \textbf{Causality} & \textbf{Crucial} & \textbf{Entity} & \textbf{Operator} & \textbf{Average} \\
& & \textbf{Inversion} & \textbf{Omission} & \textbf{Swap} & \textbf{Reversal} & \\
\midrule
\multirow{3}{*}{\textbf{Large}} 
& w/o feedback & 77.4\% & 80.2\% & 72.0\% & 78.9\% & 77.1\% \\
& w/ feedback & 98.0\% & 96.5\% & 96.0\% & 97.0\% & 96.9\% \\
&  & \textit{$\Delta$+20.6\%} & \textit{$\Delta$+16.3\%} & \textit{$\Delta$+24.0\%} & \textit{$\Delta$+18.1\%} & \textit{$\Delta$+19.8\%} \\
\midrule
\multirow{3}{*}{\textbf{Small}} 
& w/o feedback & 43.1\% & 40.8\% & 38.7\% & 39.1\% & 40.4\% \\
& w/ feedback & 72.7\% & 76.7\% & 82.4\% & 75.1\% & 76.7\% \\
&  & \textit{$\Delta$+29.6\%} & \textit{$\Delta$+35.9\%} & \textit{$\Delta$+43.7\%} & \textit{$\Delta$+36.0\%} & \textit{$\Delta$+36.3\%} \\
\bottomrule
\end{tabular}
\caption{Percentage of improved responses successfully addressing issues. We observe a consistent positive delta between the without-feedback and with-feedback variants across the four corruption types.
}
\label{tab:feedback_addressing_full_synthetic_data}
\end{table*}

\begin{table*}[t!]
\centering
\small 
\setlength{\tabcolsep}{4pt} 
\begin{tabular}{llccccc}
\toprule
\textbf{Improver} & \textbf{Judge} & \textbf{Causality} & \textbf{Crucial} & \textbf{Entity} & \textbf{Operator} & \textbf{Average} \\
& \textbf{Preference} & \textbf{Inversion} & \textbf{Omission} & \textbf{Swap} & \textbf{Reversal} & \textbf{} \\
\midrule
\multirow{3}{*}{\textbf{Large}} 
& w. Feedback & 29.5\% & 29.6\% & 26.1\% & 26.5\% & 27.9\% \\
& w/o. Feedback & \textbf{41.2\%} & \textbf{41.7\%} & \textbf{45.0\%} & \textbf{43.4\%} & \textbf{42.8\%} \\
& Tie & 29.3\% & 28.7\% & 29.0\% & 30.1\% & 29.3\% \\
\midrule
\multirow{3}{*}{\textbf{Small}} 
& w. Feedback & \textbf{49.1\%} & \textbf{53.8\%} & \textbf{56.8\%} & \textbf{56.3\%} & \textbf{54.0\%} \\
& w/o. Feedback & 38.4\% & 36.5\% & 28.0\% & 32.6\% & 33.9\% \\
& Tie & 12.6\% & 9.8\% & 15.2\% & 11.1\% & 12.2\% \\
\bottomrule
\end{tabular}
\caption{Judge pairwise evaluation of the with-feedback vs. without-feedback variants on the full synthetic data (hard prompt + creative writing). Although the without-feedback variant fixes the corruption at a lower rate, we see a consistent preference for the without-feedback variant for the large improver. The hard prompt only results are in Fig.~\ref{tab:with_feedback_vs_whithout_feedback}}
\label{tab:with_feedback_vs_whithout_feedback_full}
\end{table*}

\section{Data Extraction}

\subsection{Corruption Prompts} \label{app:corruptions_prompt}

We prompt Gemini-3-flash-preview with the corruption prompts detailed in Figures \ref{fig:prompt_causality}, \ref{fig:prompt_omission}, \ref{fig:prompt_entity}, and \ref{fig:prompt_logic}.

\begin{figure*}[t]
\begin{promptbox}[Corruption Prompt 1: Causality Inversion]
\textbf{System Role:} You are an expert data evaluator and adversarial example generator.

\textbf{Objective:} Take the provided [Question] and [Reference Response], generate a [Corrupted Response] using a Causality Inversion, and then generate simulated User Feedback correcting the error.

\textbf{Rules:}
\begin{enumerate}
    \item Identify a cause-and-effect relationship described in the response.
    \item Swap the cause and the effect, making the claim backward. Ensure the new sentence flows naturally. Do not add disclaimers.
    \item Generate three levels of simulated user feedback. The feedback must read naturally as a human chatting with an AI. Keep the feedback short, casual, and slightly lazy, mimicking how a real user would quickly type on a phone or keyboard. Do NOT mention the "corruption process" or the "prompt".
\end{enumerate}

\textbf{Output Format:} \\
\textbf{Corrupted Response:} [The fully generated text] \\
\textbf{Edit Log:} [Explain exactly the cause and effect that were swapped] \\
\textbf{Detection Difficulty:} [Rate as LOW, MEDIUM, or HIGH based strictly on how hard it is for a general reader to spot the error, and provide a 1-sentence justification.] \\
\textbf{User Feedback (Level 1 - How to fix):} [A short, lazy reply (1-2 sentences max) pointing out the reversed causality and giving the right one (e.g., "Other way around, X causes Y.")] \\
\textbf{User Feedback (Level 2 - What is the problem):} [A brief reply pointing out it's backward without explicitly fixing it (e.g., "I think you mixed up cause and effect.", "That's backward.")] \\
\textbf{User Feedback (Level 3 - Just `you are wrong'):} [A very short, blunt reply simply stating it is wrong (e.g., "Makes no sense.", "Wrong.", "Nope.")]
\end{promptbox}
\caption{Prompt for generating causality inversion corruptions.}
\label{fig:prompt_causality}
\end{figure*}

\begin{figure*}[t]
\begin{promptbox}[Corruption Prompt 2: Crucial Omission]
\textbf{System Role:} You are an expert data evaluator and adversarial example generator.

\textbf{Objective:} Take the provided [Question] and [Reference Response], generate a [Corrupted Response] using a Crucial Omission, and then generate simulated User Feedback correcting the error.

\textbf{Rules:}
\begin{enumerate}
    \item Identify a critical detail that is absolutely necessary for the response to be correct, safe, or functional.
    \item Delete this detail entirely. Smooth over the surrounding text or code so the omission is not immediately obvious. Do not add disclaimers.
    \item Generate three levels of simulated user feedback. The feedback must read naturally as a human chatting with an AI. Keep the feedback short, casual, and slightly lazy, mimicking how a real user would quickly type on a phone or keyboard. Do NOT mention the "corruption process" or the "prompt".
\end{enumerate}

\textbf{Output Format:} \\
\textbf{Corrupted Response:} [The fully generated text] \\
\textbf{Edit Log:} [Explain exactly what crucial detail was removed] \\
\textbf{Detection Difficulty:} [Rate as LOW, MEDIUM, or HIGH based strictly on how hard it is for a general reader to spot the error, and provide a 1-sentence justification.] \\
\textbf{User Feedback (Level 1 - How to fix):} [A short, lazy reply (1-2 sentences max) noting what's missing and saying what to add (e.g., "You forgot to import json.", "Add the baking powder.")] \\
\textbf{User Feedback (Level 2 - What is the problem):} [A brief reply stating that it's incomplete without providing the missing info (e.g., "Seems like a step is missing.", "This won't compile, you forgot something.")] \\
\textbf{User Feedback (Level 3 - Just 'you are wrong'):} [A very short, blunt reply simply stating it is wrong (e.g., "Incomplete.", "Broken.", "Doesn't work.")]
\end{promptbox}
\caption{Prompt for generating crucial omission corruptions.}
\label{fig:prompt_omission}
\end{figure*}

\begin{figure*}[t]
\begin{promptbox}[Corruption Prompt 3: Entity Swap]
\textbf{System Role:} You are an expert data evaluator and adversarial example generator.

\textbf{Objective:} Take the provided [Question] and [Reference Response], generate a [Corrupted Response] using an Entity Swap, and then generate simulated User Feedback correcting the error.

\textbf{Rules:}
\begin{enumerate}
    \item Identify a key entity in the reference response (e.g., a historical figure, a location, a specific number, a coding library, or a variable name).
    \item Swap it with a related, plausible, but factually incorrect entity.
    \item Keep the rest of the sentence structure and tone completely identical. Do not add disclaimers.
    \item Generate three levels of simulated user feedback. The feedback must read naturally as a human chatting with an AI. Keep the feedback short, casual, and slightly lazy, mimicking how a real user would quickly type on a phone or keyboard. Do NOT mention the "corruption process" or the "prompt".
\end{enumerate}

\textbf{Output Format:} \\
\textbf{Corrupted Response:} [The fully generated text] \\
\textbf{Edit Log:} [Explain exactly what was swapped] \\
\textbf{Detection Difficulty:} [Rate as LOW, MEDIUM, or HIGH based strictly on how hard it is for a general reader to spot the error, and provide a 1-sentence justification.] \\
\textbf{User Feedback (Level 1 - How to fix):} [A short, lazy reply (1-2 sentences max) pointing out the error and providing the fix casually (e.g., "Wait, X is actually Y.")] \\
\textbf{User Feedback (Level 2 - What is the problem):} [A brief reply pointing out what is wrong without giving the right answer (e.g., "Are you sure about X?", "I think the location is wrong.")] \\
\textbf{User Feedback (Level 3 - Just 'you are wrong'):} [A very short, blunt reply simply stating it is wrong (e.g., "Wrong.", "Nope.", "That's incorrect.")]
\end{promptbox}
\caption{Prompt for generating entity swap corruptions.}
\label{fig:prompt_entity}
\end{figure*}

\begin{figure*}[t]
\begin{promptbox}[Corruption Prompt 4: Logic Reversal]
\textbf{System Role:} You are an expert data evaluator and adversarial example generator.

\textbf{Objective:} Take the provided [Question] and [Reference Response], generate a [Corrupted Response] using a Logic Reversal, and then generate simulated User Feedback correcting the error.

\textbf{Rules:}
\begin{enumerate}
    \item Identify a logical relationship in the reference response (e.g., a mathematical operator, a boolean state, a chronological sequence, or a conditional statement).
    \item Invert or reverse this specific logic.
    \item Ensure the text or code still reads naturally and compiles/parses conceptually. Do not add disclaimers.
    \item Generate three levels of simulated user feedback. The feedback must read naturally as a human chatting with an AI. Keep the feedback short, casual, and slightly lazy, mimicking how a real user would quickly type on a phone or keyboard. Do NOT mention the "corruption process" or the "prompt".
\end{enumerate}

\textbf{Output Format:} \\
\textbf{Corrupted Response:} [The fully generated text] \\
\textbf{Edit Log:} [Explain exactly what logic was reversed] \\
\textbf{Detection Difficulty:} [Rate as LOW, MEDIUM, or HIGH based strictly on how hard it is for a general reader to spot the error, and provide a 1-sentence justification.] \\
\textbf{User Feedback (Level 1 - How to fix):} [A short, lazy reply (1-2 sentences max) pointing out the logical error and providing the fix (e.g., "You put < instead of >", "It should be True, not False.")] \\
\textbf{User Feedback (Level 2 - What is the problem):} [A brief reply pointing out where the logic breaks without giving the fix (e.g., "The math doesn't make sense there.", "Check your if statement.")] \\
\textbf{User Feedback (Level 3 - Just 'you are wrong'):} [A very short, blunt reply simply stating it is wrong (e.g., "This doesn't work.", "Wrong logic.", "Nope.")]
\end{promptbox}
\caption{Prompt for generating logic reversal corruptions.}
\label{fig:prompt_logic}
\end{figure*}

\subsection{Naturally Occurring Feedback Extraction} \label{app:nof_extraction}

To identify and categorize user feedback within the conversations, we process full conversations using the extraction prompt detailed in Figure \ref{fig:prompt_extraction}. We then filter out all UR5 feedback instances. 

To evaluate whether the simulated user feedback contains universally actionable insights or is merely subjective, we prompt Gemini-3-flash-preview using the template detailed in Figure \ref{fig:prompt_twousers}.

\begin{figure*}[t]
\begin{promptbox}[Feedback Extraction Prompt]
You are an expert dialogue analyst. Your task is to review the \textbf{entire dialogue} provided below and extract only the User utterances that contain specific feedback regarding the Assistant's performance.

\textbf{Analysis Instructions:}
\begin{enumerate}
    \item Iterate through every User utterance in the dialogue.
    \item Determine if the utterance falls into one of the \textbf{Feedback Patterns} (UR1 - UR5) defined below.
    \item If the utterance represents \textbf{Normal Conversation} (e.g., a new question, a direct answer to a question, or neutral chat), \textbf{IGNORE it}. Do not include it in the output.
    \item Output a JSON Array containing only the identified feedback instances. If no feedback is found in the entire dialogue, output an empty array \texttt{[]}.
\end{enumerate}

\textbf{Feedback Pattern Definitions:}
\begin{itemize}
    \item \textbf{Repeat or Rephrase (UR1):} The user simply repeats or rephrases their previous request because the assistant failed to address it (e.g., "Actually, I wanted...").
    \item \textbf{Make Aware with Correction (UR2):} The user explicitly points out an error AND provides the correct information (e.g., "No, I said Tuesday, not Thursday").
    \item \textbf{Make Aware without Correction (UR3):} The user indicates something is wrong or expresses dissatisfaction but does not provide the fix (e.g., "That is incorrect," "You are wrong").
    \item \textbf{Ask for Clarification (UR4):} The user expresses doubt or asks the assistant to verify its output (e.g., "Are you sure?").
    \item \textbf{Positive Feedback (UR5):} The user explicitly confirms the assistant did a good job or thanks it (e.g., "Great, thanks," "That worked").
\end{itemize}

\textbf{Input Data:} \\
{[Insert Full Dialogue Here]}

\textbf{Output Format:} \\
Provide \textbf{only} a valid JSON list. Do not include markdown formatting.

{\ttfamily
[ \\
\hspace*{1em}\{ \\
\hspace*{2em}"User Response Pattern": "UR1", \\
\hspace*{2em}"User Response Text": "Text of the specific user response" \\
\hspace*{1em}\}, \\
\hspace*{1em}\{ \\
\hspace*{2em}"User Response Pattern": "UR5", \\
\hspace*{2em}"User Response Text": "Text of another user response" \\
\hspace*{1em}\} \\
]
}
\end{promptbox}
\caption{Prompt used to extract and categorize user feedback utterances from full conversational dialogues.}
\label{fig:prompt_extraction}
\end{figure*}

\begin{table}[htbp]
\centering
\small 
\setlength{\tabcolsep}{4pt} 
\resizebox{\columnwidth}{!}{%
\begin{tabular}{lccc}
\toprule
\multirow{2}{*}{\textbf{Improver}} & \multicolumn{3}{c}{\textbf{Judge Preference}} \\
\cmidrule(lr){2-4}
& \textbf{w. Feedback} & \textbf{w/o. Feedback} & \textbf{Tie} \\
\midrule
\textbf{Large} & 32.5\% & \textbf{55.4\%} & 12.1\% \\
\textbf{Small} & 35.0\% & \textbf{51.8\%} & 13.2\% \\
\bottomrule
\end{tabular}%
}
\caption{Judge pairwise evaluation of the with-feedback vs. without-Feedback variants, for the real-user data. We see similar trends to those of the synthetic data: although the without-feedback variants fix the corruption in a lower rate, we see a consistent preference for the without-feedback variants, here not only for the large improver but also for the small improver.}
\label{tab:judge_comparison_real_user_overall}
\end{table}

\begin{table}[htbp]
\centering
\small 
\setlength{\tabcolsep}{4pt} 
\resizebox{\columnwidth}{!}{%
\begin{tabular}{llccc}
\toprule
\multirow{2}{*}{\textbf{Improver}} & \multirow{2}{*}{\textbf{$N$}} & \multicolumn{3}{c}{\textbf{Judge Preference}} \\
\cmidrule(lr){3-5}
& & \textbf{w. Feedback} & \textbf{w/o. Feedback} & \textbf{Tie} \\
\midrule
\textbf{Large} & 215 & 34.0\% & \textbf{62.8\%} & 3.3\% \\
\textbf{Small} & 274 & \textbf{54.0\%} & 40.5\% & 5.5\% \\
\bottomrule
\end{tabular}%
}
\caption{Judge preference for the \textit{WFShouldWin} subset (where No-Feedback failed to address feedback but With-Feedback succeeded), on real user data.}
\label{tab:judge_comparison_real_user_subset}
\end{table}

\subsection{Response Improvement Prompts} \label{app:improvement_prompts}

To revise and improve the model's responses, we use two variations of an improvement prompt. Figure \ref{fig:prompt_improve_feedback} is used when specific human feedback is available, while Figure \ref{fig:prompt_improve_no_feedback} relies solely on the conversation context.

\begin{figure*}[t]
\begin{promptbox}[Improve with Human Feedback Prompt]
\textbf{Task:} Revise the model's response to better address the user's intent based on specific human feedback.

\textbf{Constraints:}
\begin{enumerate}
    \item \textbf{Contextual Seamlessness}: The improved response must be self-contained and blend perfectly into the existing conversation. It must not acknowledge it is a revision or reference the "original" response.
    \item \textbf{Temporal Integrity}: Use ONLY information that was available at the time of the original user's question. Do not use hindsight or external information not present in the context.
    \item \textbf{Encapsulation}: The final polished response MUST be wrapped in double brackets [[like this]].
\end{enumerate}

\textbf{Input Data:} \\
{[}[CONVERSATION HISTORY]{]} \\
\{conversation\_history\}

{[}[ORIGINAL RESPONSE]{]} \\
\{original\_response\}

{[}[HUMAN FEEDBACK]{]} \\
\{human\_feedback\}

\textbf{Required Output Format:}
\begin{enumerate}
    \item Analysis: Identify how the original response failed to meet the criteria or the human feedback.
    \item Comparison: Explain why the improved version is a more seamless and helpful fit.
    \item Improved Response: {[}[<Insert polished response here>]{]}
\end{enumerate}
\end{promptbox}
\caption{Prompt used to revise a model response when human feedback is provided.}
\label{fig:prompt_improve_feedback}
\end{figure*}

\begin{figure*}[t]
\begin{promptbox}[Improvement without Feedback Prompt]
\textbf{Task:} Revise the model's response to better address the user's intent within the conversation context.

\textbf{Constraints:}
\begin{enumerate}
    \item \textbf{Contextual Seamlessness}: The improved response must be self-contained and blend perfectly into the existing conversation. It must not acknowledge it is a revision or reference the "original" response.
    \item \textbf{Temporal Integrity}: Use ONLY information that was available at the time of the original user's question. Do not use hindsight or external information not present in the context.
    \item \textbf{Encapsulation}: The final polished response MUST be wrapped in double brackets [[like this]].
\end{enumerate}

\textbf{Input Data:} \\
{[}[CONVERSATION HISTORY]{]} \\
\{conversation\_history\}

{[}[ORIGINAL RESPONSE]{]} \\
\{original\_response\}

\textbf{Required Output Format:}
\begin{enumerate}
    \item Analysis: Identify specific gaps, inaccuracies, or tone shifts in the original response.
    \item Comparison: Explain why the improved version is a more seamless and helpful fit.
    \item Improved Response: {[}[<Insert polished response here>]{]}
\end{enumerate}
\end{promptbox}
\caption{Prompt used to revise a model response based purely on the conversation history, without human feedback.}
\label{fig:prompt_improve_no_feedback}
\end{figure*}

\section{Evaluation Prompts} \label{app:prompt_eval}

\subsection{Issue Resolution Prompt} 

To automatically evaluate whether the revised model responses successfully incorporated the user's feedback, we use the resolution addressing prompt detailed in Figure \ref{fig:prompt_feedback_eval}.

\subsection{Pairwise Evaluation Prompt}

To evaluate the quality of the revised responses compared to the originals, we use an LLM-as-a-judge approach with the pairwise evaluation prompt detailed in Figure \ref{fig:prompt_judge}.

\begin{figure*}[t]
\begin{promptbox}[Pairwise Evaluation Prompt]
You are an expert judge evaluating AI assistant responses.

Below is a conversation history followed by two candidate responses (Response A and Response B) to the final user message.

\textbf{Conversation History:} \\
\{conversation\_history\}

\noindent\rule{\linewidth}{0.4pt}

\textbf{Response A:} \\
\{response\_a\}

\noindent\rule{\linewidth}{0.4pt}

\textbf{Response B:} \\
\{response\_b\}

\noindent\rule{\linewidth}{0.4pt}

\textbf{Task:} \\
Evaluate which response better addresses the user's needs in the context of the conversation. Consider:
\begin{itemize}
    \item Accuracy and factual correctness
    \item Helpfulness and relevance to the user's question
    \item Clarity, coherence, and completeness
    \item Appropriate tone and style for the conversation
\end{itemize}

\textbf{Output your verdict as a JSON object and nothing else:}

{\ttfamily
\{ \\
\hspace*{1em}"reasoning": "<one or two sentences explaining your choice>", \\
\hspace*{1em}"verdict": "A" | "B" | "tie" \\
\}
}
\end{promptbox}
\caption{Prompt used by the LLM-as-a-judge to evaluate and compare two candidate model responses.}
\label{fig:prompt_judge}
\end{figure*}

\begin{figure*}[t]
\begin{promptbox}[Issue Resolution Prompt]
You are an automated Quality Assurance system evaluating whether a revised AI response successfully incorporates specific user feedback.

\textbf{Conversation History (including the original response):} \\
\{conversation\_history\}

\noindent\rule{\linewidth}{0.4pt}

\textbf{User Feedback:} \\
\{feedback\}

\noindent\rule{\linewidth}{0.4pt}

\textbf{Revised Response:} \\
\{revised\_response\}

\noindent\rule{\linewidth}{0.4pt}

\textbf{Task:} \\
The user provided feedback after seeing the original response above. Evaluate whether the revised response adequately addresses that feedback. Consider:
\begin{itemize}
    \item \textbf{Intent matching}: Does the revised response specifically tackle what the feedback asked for?
    \item \textbf{Completeness}: Is the feedback fully addressed, or only partially?
    \item \textbf{Regression}: Did the revision introduce errors or hallucinate content not in the original (unless the feedback requested new content)?
\end{itemize}

\textbf{Scoring rubric:}
\begin{itemize}
    \item \textbf{1 (Fail)}: Feedback completely ignored or the revision is irrelevant to it.
    \item \textbf{2 (Poor)}: Attempted to address feedback but failed significantly (wrong direction, factual error).
    \item \textbf{3 (Partial)}: Addressed some of the feedback but missed key parts.
    \item \textbf{4 (Good)}: Addressed feedback well, with minor gaps.
    \item \textbf{5 (Perfect)}: Fully and correctly implemented the feedback without degrading quality.
\end{itemize}

\textbf{Output your verdict as a JSON object and nothing else:}

{\ttfamily
\{ \\
\hspace*{1em}"feedback\_intent": "<one sentence summarizing what the user wanted changed>", \\
\hspace*{1em}"change\_analysis": "<step-by-step comparison of original vs. revised response with respect to the feedback>", \\
\hspace*{1em}"regression\_check": "<did the revision break anything else? Yes/No + brief explanation>", \\
\hspace*{1em}"score": <1-5>, \\
\hspace*{1em}"addresses\_feedback": <true or false> \\
\}
}
\end{promptbox}
\caption{Prompt used by the issue resolution evaluator to score how well revised responses address the user feedback.}
\label{fig:prompt_feedback_eval}
\end{figure*}

\begin{figure*}[t]
\begin{promptbox}[Feedback Relevance Prompt]
\textbf{System Role:} You are an AI evaluator. Your task is to analyze user feedback on an AI response to determine if that feedback contains actionable insights that should apply to all future users asking the same question.

\textbf{The Data:}

User Question and AI response: \\
\{conversation\}

User Feedback: \{feedback\}

Is the feedback relevant to improve the response to another user asking the same question?

\textbf{Definitions:}

\textbf{[[relevant]]:} The feedback identifies factual errors, logical flaws, safety issues, or missing crucial information. Correcting this would improve the answer for anyone.

\textbf{[[irrelevant]]:} The feedback is purely subjective (e.g., "I don't like your tone," "Make it shorter") or specific to that user's unique context, and applying it might hurt the experience for a general user.

\textbf{Instructions:}

Analyze the feedback in relation to the question and response.

Determine if the feedback is objective/universal or subjective/personal.

Provide a brief explanation of your reasoning.

End with the final tag: \\
\textbf{Final Label:} [[tag]]
\end{promptbox}
\caption{Prompt used to determine if user feedback contains universally actionable insights.}
\label{fig:prompt_twousers}
\end{figure*}

\section{Improvement Type Prompt} \label{app:improvement_type}
To categorize the differences between the with-feedback vs. without-feedback improvements, we use the prompt detailed in Figure \ref{fig:prompt_improvement_type}.

\begin{figure*}[t]
\begin{promptbox}[Prompt 11: Improvement Type Evaluation]
You are an expert evaluator analyzing the differences between an original AI response and an improved version.

\textbf{Conversation History:} \\
\{conversation\_history\}

\noindent\rule{\linewidth}{0.4pt}

\textbf{Original Response:} \\
\{original\_response\}

\noindent\rule{\linewidth}{0.4pt}

\textbf{Improved Response:} \\
\{improved\_response\}

\noindent\rule{\linewidth}{0.4pt}

\textbf{Task:} \\
Compare the original and improved responses. Identify the primary type of improvement made (if any).

\textbf{Categories:}
\begin{itemize}
    \item \textbf{Content: Factuality} — The improved response corrects factual errors or inaccuracies.
    \item \textbf{Content: Completeness} — The improved response adds missing information or addresses parts of the query the original missed.
    \item \textbf{Content: Logic} — The improved response fixes logical errors, inconsistencies, or poor reasoning.
    \item \textbf{Style: Tone} — The improved response uses more appropriate tone or language for the context.
    \item \textbf{Style: Conciseness} — The improved response is more concise without losing important information.
    \item \textbf{Style: Formatting and Structure} — The improved response is better organized, uses better formatting, headers, or structure.
    \item \textbf{No Improvement} — The responses are essentially equivalent, or the "improved" version is not actually better.
\end{itemize}

\textbf{Output your verdict as a JSON object and nothing else:}

{\ttfamily
\{ \\
\hspace*{1em}"improvement\_category": "Select one: [Content: Factuality, Content: Completeness, Content: Logic, Style: Tone, Style: Conciseness, Style: Formatting and Structure, No Improvement]", \\
\hspace*{1em}"rationale": "<brief description of the specific gap in the original and how the improved version addresses it>" \\
\}
}
\end{promptbox}
\caption{Prompt used to categorize the specific type of improvement made when revising a model response.}
\label{fig:prompt_improvement_type}
\end{figure*}

\section{Human Annotation} \label{app:human_annotation}
To confirm the reliability of our automated pipelines, we conducted a human evaluation of the data extraction and evaluation processes. 

The tasks were performed by two in-house annotators: an experienced annotator and a specialist with expertise in code and mathematics. Due to the highly diverse nature of the data, achieving comprehensive coverage was unfeasible, and certain examples were consequently labeled as ‘unfamiliar’. Furthermore, owing to the broad range of domains and the granular level of detail, the annotators reported they could not conclusively guarantee that no regressions were introduced.

\subsection{Corruption Verification Guidelines}

To ensure consistent quality in the human evaluation of our corrupted dataset, annotators were provided with the instructions detailed in Figure \ref{fig:annotator_corruption}.

\subsection{Relevance Verification Guidelines} 

To evaluate the performance of our evaluator in distinguishing universal from subjective feedback, human annotators were provided with the instructions detailed in Figure \ref{fig:annotator_relevance}.

\subsection{Feedback Addressing Guidelines} 

To evaluate whether the revised model responses successfully incorporated the user feedback, human annotators were provided with the instructions detailed in Figure \ref{fig:annotator_feedback_addressing}.

\begin{figure*}[t]
\begin{tcolorbox}[
    colback=blue!5!white,
    colframe=blue!60!black,
    fonttitle=\bfseries,
    title=ANNOTATOR GUIDELINES: CORRUPTION VERIFICATION,
    arc=4pt,
    boxrule=1pt,
    left=10pt, right=10pt, top=8pt, bottom=8pt
]
\small 

You will be reviewing AI-generated responses that have been deliberately corrupted with a specific error. Your job is to verify whether the corruption actually makes the response wrong or misleading — i.e., whether the corruption "succeeded."

\vspace{0.5em}
\textbf{WHAT YOU WILL SEE PER ITEM} \\
Each item presents four fields:
\begin{itemize}
    \setlength\itemsep{0em} 
    \item \textbf{User Question}: The original question the AI was asked to answer.
    \item \textbf{Original Response}: The correct, unmodified AI response.
    \item \textbf{Corrupted Response}: The response after a deliberate error was introduced.
    \item \textbf{Edit Log}: A short description of exactly what was changed and why it is an error.
\end{itemize}

\vspace{0.5em}
\textbf{THE FOUR CORRUPTION TYPES} \\
Each item belongs to exactly one corruption type (visible in the filename). Understanding the type helps you know what to look for, but there is no need to verify the corruption type.
\begin{enumerate}
    \setlength\itemsep{0em}
    \item \textbf{Causality Inversion}: A cause-and-effect relationship was reversed. (\textit{Example:} "She failed the exam, which is why she didn't study.")
    \item \textbf{Crucial Omission}: A critical detail was removed. (\textit{Example:} Removing a required import statement from a code snippet.)
    \item \textbf{Entity / Subject Swap}: A key entity was replaced with a related but incorrect one. (\textit{Example:} A historical figure replaced with a plausible but wrong person.)
    \item \textbf{Logic Operator Reversal}: A logical relationship was inverted. (\textit{Example:} A loop condition \texttt{i < n} changed to \texttt{i > n}.)
\end{enumerate}

\vspace{0.5em}
\textbf{YOUR ANNOTATION TASK} \\
For each item, make one judgment: \textbf{Is the corrupted response wrong?} \\
Ask yourself: \textit{If a user received the corrupted response (without seeing the original), would it give them incorrect information or lead them to a wrong outcome?}
\begin{itemize}
    \setlength\itemsep{0em}
    \item \textbf{Yes}: The corrupted response contains a clear error that would mislead or fail the user.
    \item \textbf{No}: The corrupted response is still acceptable — the change is too minor, or the error doesn't actually affect the outcome.
    \item \textbf{Borderline}: There is a real error, but whether it is "clearly wrong" is genuinely debatable.
\end{itemize}

\textbf{Note on Factual Verification:} \\
If the corruption affects factual accuracy in a domain you are unfamiliar with, the edit log alone is sufficient if it convinces you. If it doesn't, you may look up the claim using external resources. You may use an LLM to help locate a relevant source, but not to judge correctness directly. If you still cannot verify it, label "\textbf{Unfamiliar}".

\vspace{0.5em}
\textbf{SPECIAL NOTES FOR CREATIVE WRITING ITEMS}
\begin{itemize}
    \setlength\itemsep{0em}
    \item Check that the corrupted response still fulfills the user question's requirements.
    \item Focus on internal consistency: does the corrupted version contradict itself or reverse a story's logic?
    \item A causality inversion in a motivational story is a real error if the story's message is broken.
    \item Omitting a character trait or detail that the rest of the story depends on counts as a valid corruption (\textbf{'Yes'} label).
\end{itemize}
\end{tcolorbox}
\caption{Guidelines provided to human annotators for verifying the validity and severity of the generated corruptions.}
\label{fig:annotator_corruption}
\end{figure*}

\begin{figure*}[t]
\begin{tcolorbox}[
    colback=blue!5!white,
    colframe=blue!60!black,
    fonttitle=\bfseries,
    title=ANNOTATOR GUIDELINES: RELEVANCE VERIFICATION,
    arc=4pt,
    boxrule=1pt,
    left=10pt, right=10pt, top=8pt, bottom=8pt
]
\small

You will be reviewing a user's question, an AI's original response, a user feedback that critiques that response, and an AI evaluator's assessment of whether that feedback is universally applicable. Your job is to evaluate whether the AI evaluator correctly tagged the feedback as relevant or irrelevant based on its reasoning and the provided context.

\vspace{0.5em}
\textbf{WHAT YOU WILL SEE PER ITEM}
\begin{itemize}
    \setlength\itemsep{0em}
    \item \textbf{Question}: The original prompt provided by the user.
    \item \textbf{Model Response}: The AI's original answer to the user's question.
    \item \textbf{Feedback Text (User Response)}: The specific critique, correction, or request the user provided regarding the AI's answer.
    \item \textbf{Model Tag}: The AI evaluator's classification of the feedback—either \texttt{[[relevant]]} or \texttt{[[irrelevant]]}.
    \item \textbf{Model Reasoning}: The explanation for why the AI evaluator chose that specific tag.
\end{itemize}

\vspace{0.5em}
\textbf{YOUR ANNOTATION TASK} \\
For each item, make one judgment: Compare the user's feedback against the AI evaluator's tag and reasoning to determine whether the model\_tag is correct.

\vspace{0.5em}
\textbf{LABELS} \\
Your answer must be a strict \textbf{Correct} or \textbf{Incorrect}. Use the following definitions to guide your choice:

\begin{itemize}
    \setlength\itemsep{0.25em}
    \item \textbf{CORRECT} -- Select CORRECT if the AI evaluator's tag perfectly matches the true nature of the feedback, supported by sound reasoning.
    \begin{itemize}
        \setlength\itemsep{0em}
        \item \textbf{Accurate Universal Tagging}: The feedback points out a factual error, logical flaw, safety issue, or missing crucial information, and the model correctly tagged it as \texttt{[[relevant]]}. Correcting this would improve the answer for anyone asking the same question.
        \item \textbf{Accurate Subjective Tagging}: The feedback is based on personal preference (e.g., "Make it shorter," "I don't like your tone") or a unique user context, and the model correctly tagged it as \texttt{[[irrelevant]]}. Applying this change might hurt a general user's experience.
    \end{itemize}

    \vspace{0.5em}
    \item \textbf{INCORRECT} -- Select INCORRECT if the AI evaluator assigned the wrong tag or used fundamentally flawed reasoning. Apply this label if any of the following are true:
    \begin{itemize}
        \setlength\itemsep{0em}
        \item \textbf{Missed Universality}: The user pointed out an objective error (e.g., incorrect math, dangerous advice, broken code), but the model incorrectly tagged it as \texttt{[[irrelevant]]}.
        \item \textbf{False Universality}: The user asked for a purely subjective change (e.g., "Rewrite this as a poem," "I prefer bullet points"), but the model incorrectly tagged it as \texttt{[[relevant]]}.
        \item \textbf{Flawed Reasoning}: The tag happens to be right, but the model's reasoning shows a complete misunderstanding of the feedback or the conversation context.
    \end{itemize}
\end{itemize}

\end{tcolorbox}
\caption{Guidelines provided to human annotators for verifying the evaluator's classification of user feedback relevance.}
\label{fig:annotator_relevance}
\end{figure*}

\begin{figure*}[t]
\begin{tcolorbox}[
    colback=blue!5!white,
    colframe=blue!60!black,
    fonttitle=\bfseries,
    title=ANNOTATOR GUIDELINES: FEEDBACK ADDRESSING,
    arc=4pt,
    boxrule=1pt,
    left=10pt, right=10pt, top=8pt, bottom=8pt
]
\small

You will be reviewing a conversation of a user with an AI, a user feedback that critiques or corrects the last AI response, and the following AI's revised response. Your job is to evaluate whether the AI's revised response successfully resolves the user feedback.

\vspace{0.5em}
\textbf{WHAT YOU WILL SEE PER ITEM}
\begin{itemize}
    \setlength\itemsep{0em}
    \item \textbf{Conversation History}: The context of the interaction leading up to the issue.
    \item \textbf{User Feedback}: The specific critique, correction, or request the user provided regarding the previous response.
    \item \textbf{Revised Response}: The new response generated to address the feedback.
\end{itemize}

\vspace{0.5em}
\textbf{YOUR ANNOTATION TASK} \\
For each item, make one judgment: Compare the revision against the feedback to determine whether the revision addressed the user feedback or not.

\vspace{0.5em}
\textbf{LABELS} \\
Your answer must be a strict \textbf{Yes} or \textbf{No}. Use the following definitions to guide your choice:

\begin{itemize}
    \setlength\itemsep{0.25em}
    \item \textbf{YES} -- Select YES if the revised response specifically and successfully tackles the core of what the feedback asked for.
    \begin{itemize}
        \setlength\itemsep{0em}
        \item \textbf{The requirement is met}: If the user asked for a bulleted list, the response is a bulleted list.
        \item \textbf{The correction is made}: If the user pointed out a factual error, the error is removed or corrected.
        \item \textbf{Spirit of the prompt}: The revision maintains the original context while seamlessly integrating the requested change.
    \end{itemize}

    \vspace{0.5em}
    \item \textbf{NO} -- Select NO if the revised response fails to meet the user's demand. Apply this label if any of the following are true:
    \begin{itemize}
        \setlength\itemsep{0em}
        \item \textbf{Ignored}: The response completely ignores the feedback.
        \item \textbf{Partial Compliance}: The user asked for two specific things (e.g., "Make it shorter and add a Spanish translation"), but the revision only accomplished one.
        \item \textbf{Superficial Fix}: The response acknowledges the feedback in text (e.g., "I have updated the tone to be more formal") but the actual content remains unchanged or fails to achieve the requested result.
        \item \textbf{Regression}: The revision fixes the feedback but severely breaks another critical part of the original prompt/context (e.g., the tone is fixed, but the answer is suddenly entirely incorrect).
    \end{itemize}
\end{itemize}

\end{tcolorbox}
\caption{Guidelines provided to human annotators for evaluating whether a revised AI response successfully addressed the user's feedback.}
\label{fig:annotator_feedback_addressing}
\end{figure*}

\begin{table*}[htbp]
\centering
\small 
\setlength{\tabcolsep}{4pt} 
\begin{tabular}{llccccc}
\toprule
\textbf{Improver} & \textbf{Variant} & \textbf{Causality} & \textbf{Crucial} & \textbf{Entity} & \textbf{Operator} & \textbf{Average} \\
& & \textbf{Inversion} & \textbf{Omission} & \textbf{Swap} & \textbf{Reversal} & \\
\midrule
\multirow{3}{*}{\textbf{Gemini-3-Flash}} 
& With-Feedback & 91.1\% & 94.6\% & 84.4\% & 93.5\% & 90.9\% \\
& No-Feedback & 84.8\% & 84.0\% & 75.0\% & 83.3\% & 81.7\% \\
&  & \textit{$\Delta$+6.3\%} & \textit{$\Delta$+10.6\%} & \textit{$\Delta$+9.4\%} & \textit{$\Delta$+10.2\%} & \textit{$\Delta$+9.2\%} \\
\bottomrule
\end{tabular}
\caption{Percentage of Responses Successfully Addressing Feedback (Level 2)}
\label{tab:feedback_addressing_level2}
\end{table*}

\begin{table*}[htbp]
\centering
\small 
\setlength{\tabcolsep}{4pt} 
\begin{tabular}{llccccc}
\toprule
\textbf{Improver} & \textbf{Variant} & \textbf{Causality} & \textbf{Crucial} & \textbf{Entity} & \textbf{Operator} & \textbf{Average} \\
& & \textbf{Inversion} & \textbf{Omission} & \textbf{Swap} & \textbf{Reversal} & \\
\midrule
\multirow{3}{*}{\textbf{Gemini-3-Flash}} 
& With-Feedback & 79.6\% & 90.2\% & 80.6\% & 89.4\% & 84.9\% \\
& No-Feedback & 84.8\% & 84.0\% & 75.0\% & 83.3\% & 81.7\% \\
&  & \textit{$\Delta$-5.2\%} & \textit{$\Delta$+6.2\%} & \textit{$\Delta$+5.6\%} & \textit{$\Delta$+6.1\%} & \textit{$\Delta$+3.2\%} \\
\bottomrule
\end{tabular}
\caption{Percentage of Responses Successfully Addressing Feedback (Level 3)}
\label{tab:feedback_addressing_level3}
\end{table*}

\begin{table*}[t!]
\centering
\small 
\setlength{\tabcolsep}{4pt} 
\begin{tabular}{llcccccc}
\toprule
\textbf{Improver} & \textbf{Judge} & \textbf{Causality} & \textbf{Crucial} & \textbf{Entity} & \textbf{Operator} & \textbf{Average} & \textbf{w. Feedback} \\
& \textbf{Preference} & \textbf{Inversion} & \textbf{Omission} & \textbf{Swap} & \textbf{Reversal} & \textbf{} & \textbf{Should Win} \\
\midrule
\multirow{3}{*}{\textbf{Large}} 
& w. Feedback & 28.5\% & 28.7\% & 25.9\% & 25.8\% & 27.2\% & \textbf{54.6\%} ($N$=205) \\
& w/o. Feedback & \textbf{36.9\%} & \textbf{37.5\%} & \textbf{38.7\%} & \textbf{37.9\%} & \textbf{37.8\%} & 39.0\% \\
& Tie & 34.6\% & 33.8\% & 35.4\% & 36.3\% & 35.0\% & 6.3\% \\
\midrule
\multirow{3}{*}{\textbf{Small}} 
& w. Feedback & \textbf{45.7\%} & \textbf{50.0\%} & \textbf{52.4\%} & \textbf{56.4\%} & \textbf{51.2\%} & \textbf{80.8\%} ($N$=574) \\
& w/o. Feedback & 39.2\% & 37.4\% & 30.0\% & 31.2\% & 34.3\% & 17.3\% \\
& Tie & 15.1\% & 12.6\% & 17.7\% & 12.5\% & 14.5\% & 1.9\% \\
\bottomrule
\end{tabular}
\caption{Judge pairwise evaluation of the with-feedback vs. without-Feedback variants for the synthetic data.}
\label{tab:combined_feedback}
\end{table*}

\begin{table}[htbp]
\centering
\small 
\setlength{\tabcolsep}{4pt} 
\resizebox{\columnwidth}{!}{%
\begin{tabular}{lccc}
\toprule
\multirow{2}{*}{\textbf{Subset Group}} & \multicolumn{3}{c}{\textbf{Judge Preference}} \\
\cmidrule(lr){2-4}
& \textbf{w. Feedback} & \textbf{w/o. Feedback} & \textbf{Tie} \\
\midrule
\ Independent Success    & \textbf{87.7\%} & 12.3\% & 0\% \\
\ Feedback Dependent & 72.2\% & 27.8\% & 0\% \\
\bottomrule
\end{tabular}%
}
\caption{Judgments Distribution: Group A vs. Group B}
\label{tab:judgements_distribution}
\end{table}

\begin{table}[htbp]
\centering
\small 
\setlength{\tabcolsep}{4pt} 
\resizebox{\columnwidth}{!}{%
\begin{tabular}{lccc}
\toprule
\multirow{3}{*}{\textbf{Baseline}} & \multicolumn{3}{c}{\textbf{Judge Preference}} \\
\cmidrule(lr){2-4}
& \begin{tabular}{@{}c@{}}\textbf{Reimprove} \\ \textbf{w. Feedback}\end{tabular} & \textbf{Baseline} & \textbf{Tie} \\
\midrule
\textbf{w/o. Feedback} 
& \textbf{51.0\%} & 37.6\% & 11.8\% \\
\midrule
\textbf{Reimprove w/o. Feedback} 
& 28.9\% & \textbf{57.3\%} & 13.7\% \\
\bottomrule
\end{tabular}%
}
\caption{Reimprovements results. The judge prefers the two stage with-feedback improvement over the single stage baseline, but when comparing against two stage without-feedback its reference reversed.}
\label{tab:reimprove}
\end{table}

\section{Re-improvements} \label{app:reimprove}

Another direction we explore is iterative improvement: applying an initial improvement step without feedback, followed by a second step that incorporates feedback. This sequential approach has the potential to combine the best of both worlds, merging the targeted corrections of the with-feedback variant with the broader stylistic enhancements of the without-feedback variant (see \S\ref{sec:improvement_type}).

We run this analysis for the large improver only. We isolate the subset of queries where the initial with-feedback response ($r_{+f}$) successfully addressed the feedback, but the no-feedback response ($r_{-f}$) did not ($WFShouldWin(M_i^{large})$).  Because $r_{-f}$ remains corrupted, we know it requires further refinement. Furthermore, since $r_{+f}$ successfully resolved the issue for these specific queries, there is strong reason to believe that a secondary improvement step utilizing feedback will be effective.

We take $r_{-f}$ and apply a second improvement step—this time using feedback—yielding $M_i(q, r_{-f}, f)$. We first evaluate this new output against the single-step $r_{-f}$. To ensure a fair comparison regarding the computational resources used, we also introduce a secondary baseline where $r_{-f}$ is improved a second time \textit{without} feedback, denoted as $M_i(q, r_{-f})$.

Table~\ref{tab:reimprove} shows that the judge prefers the two-stage with-feedback response ($M_i(q, r_{-f}, f)$) over the single-stage baseline ($r_{-f}$). However, when comparing the two two-stage methods against each other, the judge's preference reverses, favoring the response improved twice \textit{without} feedback ($M_i(q, r_{-f})$). Even when examining only the subset of cases ($n=123$) where the two-stage with-feedback approach successfully addressed the feedback but the two-stage without-feedback approach failed to do so, the judge correctly prefers the with-feedback variant in only $34.2\%$ of the cases.

It seems that the judge is heavily biased toward the stylistic polish of the without-feedback variant, often prioritizing general fluency and formatting over strict adherence to the provided feedback.

\section{Initial Results with Another Improver} \label{app:gpt_oss}

We present initial results with \textit{GPT-OSS-20B} \citep{agarwal2025gpt} on about $400$ synthetic data samples, see tables \ref{tab:feedback_addressing_gpt} and \ref{tab:with_feedback_vs_whithout_feedback_gpt_oss}.

\begin{table}[htbp]
\centering
\small 
\begin{tabular}{llc}
\toprule
\textbf{Improver} & \textbf{Variant} & \textbf{Average} \\
\midrule
\multirow{3}{*}{\textbf{GPT-OSS}} 
& w/o feedback & 48.5\% \\
& w/ feedback & 54.7\% \\
&  & \textit{$\Delta$+6.2\%} \\
\bottomrule
\end{tabular}
\caption{Percentage of improved responses successfully addressing the issues on the hard prompt category with GPT-OSS as the improver. We observe a consistent positive delta between the without-feedback and with-feedback variants.
}
\label{tab:feedback_addressing_gpt}
\end{table}

\begin{table}[htbp]
\centering
\small 
\begin{tabular}{llcc}
\toprule
\textbf{Improver} & \textbf{Judge Preference} & \textbf{Average} & \textbf{WFShouldWin} \\
\midrule
\multirow{3}{*}{\textbf{GPT-OSS}} 
& w. Feedback & \textbf{42.9\%} & \textbf{95.2\%} \\
& w/o. Feedback & 40.6\% & 4.8\% \\
& Tie & 16.5\% & 0.0\% \\
\bottomrule
\end{tabular}
\caption{Judge pairwise evaluation of the with-feedback vs. without-feedback variants for the openai\_gpt-oss-20b improver, averaged across corruption types. The `WFShouldWin` column represents the subset of cases where the with-feedback variant successfully addressed the issue while the without-feedback variant did not.}
\label{tab:with_feedback_vs_whithout_feedback_gpt_oss}
\end{table}

\section{Computational Budget} \label{app:budget}

We spent about $2,000\$$ on the Gemini API (generating corruptions, improvements, relevance evaluation, and judgments).

To run Qwen3-8B we used a 45G GPU (a40 or the like). Generating improvements for $100$ samples took about 2.5 hours, so the total generations for the main experiments took about $75$ hours. Generating judgements for the self-judge experiments took another $35$ hours.

To generate the initial results for GPT-OSS-20B (\S\ref{app:gpt_oss}) we used a h200, for a total of $10$ hours.

\end{document}